\documentclass[11pt,a4paper]{article}
\usepackage[hyperref]{acl2021}
\usepackage{times}
\usepackage{latexsym}

\usepackage{microtype}
\usepackage{tikz-dependency}
\usepackage{booktabs}
\usepackage{multirow}
\usepackage{xspace}
\usepackage{xcolor}
\usepackage{hyperref}
\usepackage{graphicx}
\usepackage{subcaption}
\usepackage{relsize}
\usepackage{amsmath}

\aclfinalcopy 
\def\aclpaperid{2862} 

\definecolor{lightblue}{RGB}{205,210,255}
\definecolor{lightred}{RGB}{255,240,240}

\newcommand{\impr}[1]{\colorbox{lightblue}{#1}}

\newcommand{\ie}{\textit{i.e.}\xspace}

\newcommand{\eg}{\textit{e.g.}\xspace}
\newcommand{\F}{F$_1$\xspace}
\newcommand{\norec}{\textbf{NoReC}$_{\text{Fine}}$\xspace}
\newcommand{\cat}{\textbf{MultiB}$_{\text{CA}}$\xspace}
\newcommand{\basque}{\textbf{MultiB}$_{\text{EU}}$\xspace}

\newcommand{\dsu}{\textbf{DS}$_{\text{Unis}}$\xspace}
\newcommand{\mpqa}{\textbf{MPQA}\xspace}

\title{Structured Sentiment Analysis as Dependency Graph Parsing}

\author{Jeremy Barnes\textsuperscript{*}, Robin Kurtz\textsuperscript{$\dagger$},  Stephan Oepen\textsuperscript{*}, Lilja Øvrelid\textsuperscript{*}and Erik Velldal\textsuperscript{*}\\[1ex]
  \textsuperscript{*}University of Oslo,  Department of Informatics\\[1ex]
  \textsuperscript{$\dagger$}National Library of Sweden, KBLab\\
  $\{\,$\texttt{jeremycb}$\,|\,$\texttt{oe}$\,|\,$\texttt{liljao}$\,|\,$\texttt{erikve}$\,\}\,$\texttt{@ifi.uio.no}\\ \texttt{robin.kurtz@kb.se}}
\date{}

\begin{document}
\maketitle
\begin{abstract}

Structured sentiment analysis attempts to extract full opinion tuples from a text, but over time this task has been subdivided into smaller and smaller sub-tasks, \eg,\ target extraction or targeted polarity classification. 
We argue that this division has become counterproductive and propose a new unified framework to remedy the situation. We cast the structured sentiment problem as dependency graph parsing, where the nodes are spans of sentiment holders, targets and expressions, and the arcs are the relations between them. We perform experiments on five datasets in four languages (English, Norwegian, Basque, and Catalan) and show that this approach leads to strong improvements over state-of-the-art baselines. Our analysis shows that refining the sentiment graphs with syntactic dependency information further improves results.

\end{abstract}

\section{Introduction}

\begin{figure*}[]
\centering
\includegraphics[width=.85\textwidth]{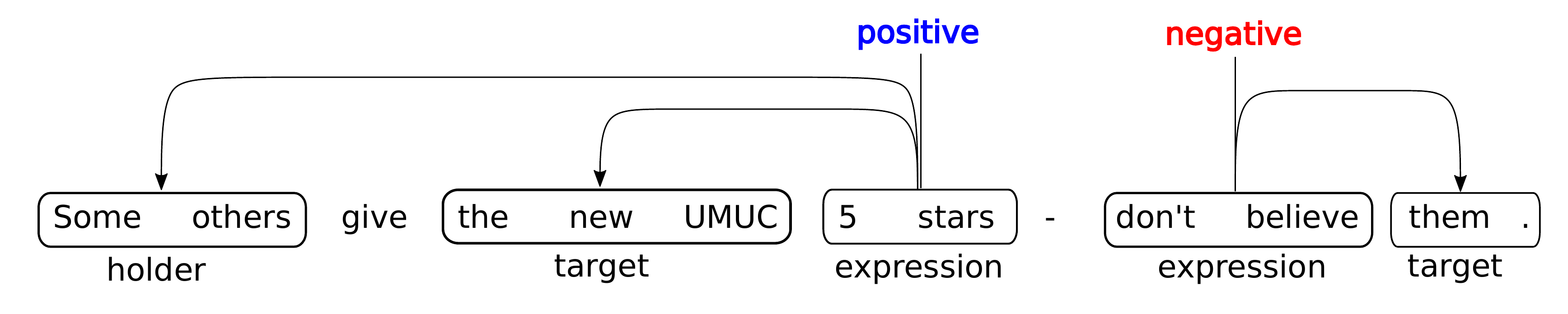}
   
   
   
   
\caption{A structured sentiment graph is composed of a holder, target, sentiment expression, their relationships and a polarity attribute. Holders and targets can be null.}
\label{deptree}
\end{figure*}

\textbf{Structured}\footnote{We use the term `structured sentiment' distinctly from \newcite{structuredsent}, who use it to refer to the latent hierarchical structure of sentiment aspects. We instead use `structured' to refer to predicting sentiment graphs as a structured prediction task, as opposed to the many text classification task that are found in sentiment analysis.} \textbf{sentiment analysis}, \ie, the task of predicting a structured sentiment graph like the ones in Figure \ref{deptree}, can be theoretically cast as an information extraction problem in which one attempts to find all of the opinion tuples $O = O_i,\ldots,O_n$ in a text. Each opinion $O_i$ is a tuple $(h, t, e, p)$ 
%
%
where $h$ is a \textbf {holder} who expresses a \textbf{polarity} $p$ towards a \textbf{target} $t$ through a \textbf{sentiment expression} $e$, implicitly defining pairwise relationships between elements of the same tuple. \newcite{Liu2012-saom} argues that all of these elements\footnote{\newcite{Liu2012-saom}'s definition replaces sentiment expression with the time when the opinion was expressed.} are essential to fully resolve the sentiment analysis problem. However, most research on sentiment analysis focuses either on a variety of sub-tasks, which avoids performing the full task, or on simplified and idealized tasks, \eg,\ sentence-level binary polarity classification. 

We argue that the division of structured sentiment into these sub-tasks has become counterproductive, as reported experiments are often not sensitive to whether a given addition to the pipeline improves the overall resolution of sentiment, or do not take into account the inter-dependencies of the various sub-tasks.  
As such, we propose a unified approach to \emph{structured sentiment} which jointly predicts all elements of an opinion tuple and their relations. Moreover,  we cast sentiment analysis as a \emph{dependency graph parsing problem}, where the sentiment expression is the root node, and the other elements have arcs which model the relationships between them. This methodology also enables us to take advantage of recent improvements in semantic dependency parsing \cite{dozat-manning-2018-simpler,oepen2020mrp,kurtz-etal-2020-end} to efficiently learn a sentiment graph parser. 

This perspective also allows us to unify a number of approaches, including targeted, and opinion tuple mining. We aim to answer \textbf{RQ1:}\ whether graph-based approaches to structured sentiment outperform state-of-the-art sequence labeling approaches, and \textbf{RQ2:}\ how to best encode structured sentiment as parsing graphs.
We perform experiments on five standard datasets in four languages (English, Norwegian, Basque, Catalan) and show that graph-based approaches outperform state-of-the-art baselines on all datasets on several standard metrics, as well as our proposed novel (unlabeled and labeled) sentiment graph metrics. We further propose methods to inject linguistic structure into the sentiment graphs using syntactic dependencies. Our main contributions are therefore 1) proposing a holistic approach to structured sentiment through sentiment graph parsing, 2) introducing new evaluation metrics for measuring model performance, and 3) extensive experimental results that outperform state-of-the-art baselines. Finally, we release the code and datasets\footnote{Code and datasets available at \url{https://github.com/jerbarnes/sentiment_graphs}.} to enable future work on this problem.

\section{Related Work}

\paragraph{\emph{Structured sentiment analysis}} can be broken down into five sub-tasks: i) sentiment expression extraction, ii) sentiment target extraction, iii) sentiment holder extraction, iv) defining the relationship between these elements, and v) assigning polarity.
Previous work on information extraction has used pipeline methods which first extract the holders, targets, and expressions (tasks \textit{i} - \textit{iii}) and subsequently predict their relations (task \textit{iv}), mostly on the \mpqa dataset \cite{Wiebe2005}. CRFs and a number of external resources (sentiment lexicons, dependency parsers, named-entity taggers) \cite{choi-etal-2006-joint,yang-cardie-2012-extracting} are strong baselines. Given the small size of the training data and the complicated task, these techniques often still outperform neural models, such as BiLSTMs \cite{katiyar-cardie-2016-investigating}. Transition-based end-to-end approaches have shown some potential \cite{ZHANG201956}. However, all of this work ignores the polarity classification subtask.

\emph{Targeted sentiment analysis} only concentrates on extracting sentiment targets (task \textit{ii}) and classifying the polarity directed towards them (task \textit{iv}) \cite{jiang-etal-2011-target,mitchell-etal-2013-open}. Recent shared tasks on \emph{Aspect-Based Sentiment Analysis} (ABSA) \cite{pontiki-etal-2014-semeval,pontiki-etal-2015-semeval,pontiki-etal-2016-semeval} also include target extraction and polarity classification subtasks. Joint approaches perform on par with pipeline methods \cite{li2019unified} and multitask models can perform even better \cite{he-etal-2019-interactive}. Finally, pretrained language models \cite{devlin-etal-2019-bert} can also lead to improvements on the ABSA data \cite{li-etal-2019-exploiting}.

\emph{End2End} sentiment analysis is a recently proposed subtask which combines targeted sentiment (tasks \textit{ii} and \textit{v}) and sentiment expression extraction (task \textit{i}), without requiring the resolution of relationships between targets and expressions. \newcite{wang-etal-2016-recursive} augment the ABSA datasets with sentiment expressions, but provide no details on the annotation process or any inter-annotator agreement. \newcite{he-etal-2019-interactive} make use of this data and propose a multi-layer CNN (\textbf{IMN}) to create hidden representations $h$ which are then fed to a target and opinion extraction module (AE), which is also a multi-layer CNN. This module predicts $\hat{y}^{ae}$, a sequence of BIO tags\footnote{The tags include \texttt{\{BIO\}}-\texttt{\{target,expression\}}} that predict the presence or absence of targets and expressions. After jointly predicting the targets and expressions, a second multi-layer CNN with a final self-attention network is used to classify the polarity, again as sequence labeling task (AS). This second module combines the information from $h$ and $\hat{y}^{ae}$ by incorporating the predicted probability of a token to be a target in the formulation of self-attention. Finally, an iterative message-passing algorithm updates $h$ using the predictions from all the modules at the previous timestep.

\newcite{chen-qian-2020-relation} instead propose Relation-Aware Collaborative Learning (\textbf{RACL}). This model creates task specific representations by first embedding a sentence, passing through a shared feed-forward network and finally a task-specific CNN. This approach then models interactions between each pair of sub-tasks (target extraction, expression extraction, sentiment classification) by creating pairwise weighted attention representations. These are then concatenated and used to create the task-specific predictions. The authors finally stack several RACL layers, using the output from the previous layer as input for the next.

Both models perform well on the augmented SemEval data, but it is unlikely that these annotations are adequate for full structured sentiment, as \newcite{wang-etal-2016-recursive} only provide expression annotations for sentences that have targets, generally only include sentiment-bearing words (not phrases), and do not specify the relationship between target and expression.

Finally, the recently proposed \emph{aspect sentiment triplet extraction} \cite{peng2019knowing,xu-etal-2020-position} attempts to extract targets, expressions and their polarity. However, the datasets used are unlikely to be adequate, as they augment available targeted datasets, but do not report annotation guidelines, procedure, or inter-annotator agreement.

\begin{table*}[t]
\newcommand{\sep}{\cmidrule(lr){3-4}\cmidrule(lr){5-7}\cmidrule(lr){8-10}\cmidrule(lr){11-13}\cmidrule(l){14-16}}
    \centering
    \resizebox{\textwidth}{!}{%
    \begin{tabular}{@{}llrrrrrrrrrrrrrr@{}}
    \toprule
      & & \multicolumn{2}{c}{sentences} & \multicolumn{3}{c}{holders} & \multicolumn{3}{c}{targets} & \multicolumn{3}{c}{expressions} & \multicolumn{3}{c}{polarity}\\
      \sep
       & & \# & avg. & \# & avg. & max & \# & avg. & max & \# & avg. & max & $+$ & neu & $-$\\
    \sep
   \norec & train & 8634 & 16.7 & 898 & 1.1 & 12 & 6778 & 1.9 & 35 & 8448 & 4.9 & 40 & 5684& 0 & 2756 \\
   & dev & 1531 & 16.9 & 120 & 1.0 & 3 & 1152 & 2.0 & 15 & 1432 & 5.1 & 31 & 988 & 0 & 443\\
   & test & 1272 & 17.2 & 110 & 1.0 & 3 & 993 & 2.0 & 20 & 1235 & 4.9 & 30 & 875 & 0 & 358\\
  \sep
     \cat & train & 1174 & 15.6 & 169 & 1.1 & 4 & 1695 & 2.4 & 18 & 1981 & 2.6 & 19 & 1272 & 0 &708 \\
   & dev & 168 & 13.3 & 15 & 1.5 & 7 & 211 & 2.3 & 10 & 258 & 2.6 & 9 & 151 & 0 &107 \\
   & test & 336 & 14.7 & 52 & 1.1 & 5 & 430 & 2.6 & 12 & 518 & 2.7 & 14 & 313 & 0 &204\\
  \sep
   \basque & train & 1064 & 10.5 & 205 & 1.1 & 6 & 1285 & 1.4 & 9 & 1684 & 2.2 & 10 & 1406 & 0 & 278\\
   & dev & 152 & 10.7 & 33 & 1.1 & 2 & 153 & 1.3 & 6 & 204 &   2.5 & 8 & 168 & 0 & 36 \\
   & test & 305 & 10.7 & 58 & 1.1 & 2 & 337 & 1.4 & 8 & 440 & 2.2 & 9 &  375 & 0 & 65\\
  \sep
  
    \mpqa       & train & 4500 & 25 & 1306 & 2.6 & 27  & 1382 & 6.1 & 56 &  1656 & 2.4 & 14  & 675 & 271 & 658 \\
                  & dev & 1622 & 23 & 377 & 2.6 & 16 & 449 & 5.3 & 41 & 552 & 2.1 & 8 & 241 & 105 & 202\\
                  & test & 1681 & 24 & 371 & 2.8 & 32 & 405 & 6.4 & 42 & 479 & 2.0 & 8 & 166 & 89 & 199\\
      \sep
       \dsu &  train & 2253 & 20 & 65 & 1.2 & 2 & 1252 & 1.2 & 5 & 837 & 1.9 & 9 & 495 & 149 & 610 \\
                  & dev & 232 & 9 & 17 & 1.1 & 3 & 151 & 1.2 & 3 & 106 & 1.7 & 6  & 40 & 19 & 92\\
                  & test & 318 & 20 & 12 & 1.3 & 4 & 198 & 1.2 & 6 & 139 & 2.0 & 5 &  77 & 18 & 103\\
                 \bottomrule
    \end{tabular}
}%
    
    \caption{Statistics of the datasets, including number of sentences and average length (in tokens) per split, as well as average and max lengths (in tokens) for holder, target, and expression annotations. Additionally, we include the distribution of polarity -- restricted to positive, neutral, and negative -- in each dataset.}
    \label{tab:statistics}
\end{table*}


\paragraph{Graph parsing:}
Syntactic dependency graphs are regularly used in applications, supplying them with necessary grammatical information \citep{mintz2009distant,cui2005question,bjorne2009extracting,johansson2012relational,lapponi2012uio2}.
The dependency graph structures used in these systems are predominantly restricted to trees.
While trees are sufficient to encode syntactic dependencies, they are not expressive enough to handle \textit{meaning representations}, that require nodes to have multiple incoming arcs, or having no incoming arcs at all \citep{kuhlmann2016catalogue}.
While much of the early research on parsing these new structures \citep{oepen2014semeval,oepen2015semeval} focused on specialized decoding algorithms, \citet{dozat-manning-2018-simpler} presented a neural dependency parser that essentially relies only on its neural network structure to predict any type of dependency graph without restrictions to certain structures.
Using the parser's ability to learn arbitrary dependency graphs, \citet{kurtz-etal-2020-end} phrased the task of negation resolution \citep{morante2012sem,morante2012conandoyleneg} as a graph parsing task.
This transformed the otherwise flat representations to dependency structures that directly encode the often overlapping relations between the building blocks of multiple negation instances at the same time.
In a simpler fashion, \citet{yu2020named} exploit the parser of \citet{dozat-manning-2018-simpler} to predict spans of named entities.

\section{Datasets}

We here focus on datasets that annotate the full task of structured sentiment as described initially.
We perform experiments on five structured sentiment datasets in four languages, the statistics of which are shown in Table~\ref{tab:statistics}. The largest available structured sentiment dataset is the \norec dataset \cite{ovrelid-etal-2020-fine}, a multi-domain dataset of professional reviews in Norwegian, annotated for structured sentiment. \basque and \cat  \cite{barnes-etal-2018-multibooked} are hotel reviews in Basque and Catalan, respectively. \mpqa \cite{Wiebe2005} annotates news wire text in English. Finally, \dsu \cite{toprak-etal-2010-sentence} annotate English reviews of online universities and e-commerce. In our experiments, we use only the university reviews, as the e-commerce reviews have a large number of `polar targets', \ie, targets with a polarity, but no accompanying sentiment expression.

While all the datasets annotate holders, targets, and expressions, the frequency and distribution of these vary. Regarding holders, \mpqa has the most (2,054) and \dsu has the fewest (94), whereas \norec has the largest proportion of targets (8,923) and expressions (11,115). The average length of holders (2.6 tokens) and targets (6.1 tokens) in \mpqa is also considerably higher than the others.

It is also worth pointing out that \mpqa and \dsu additionally include neutral polarity. In the case of \mpqa the neutral class refers to verbs which are subjective but do not convey polarity, \eg, `say', `opt for'. In \dsu, however, the neutral label tends to indicate expressions that could entail mixed polarity or are polar under the right conditions, \eg, `the classes were not easy' is considered neutral, as it is possible for difficult classes to be desirable at a university. \basque, and \cat also have labels for strong positive and strong negative, which we map to positive and negative, respectively. Finally, \norec includes intensity annotations (strong, normal, slight), which we disregard for the purposes of these experiments.

\begin{figure*}
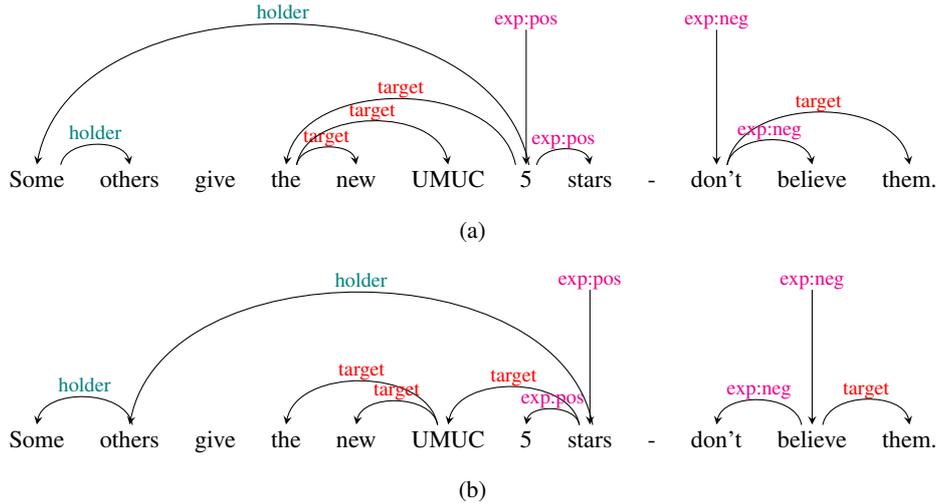

\centering
\begin{subfigure}[head-first]{0.9\textwidth}
\centering
\smaller
\begin{dependency}[theme = simple]
   \begin{deptext}[column sep=1em]
     Some \& others \& give \& the \& new \& UMUC \& 5 \& stars \& - \& don't \& believe \& them.\\
   \end{deptext}
   \deproot{7}{\color{magenta}{\normalsize{exp:pos}}}
   \deproot{10}{\color{magenta}{\normalsize{exp:neg}}}
   \depedge[]{7}{4}{\normalsize{\color{red}{target}}}
   \depedge[]{4}{5}{\normalsize{\color{red}{target}}}
   \depedge[]{4}{6}{\normalsize{\color{red}{target}}}
   \depedge{7}{8}{\normalsize{\color{magenta}{exp:pos}}}
   \depedge[edge start x offset=5pt]{7}{1}{\normalsize{\color{teal}{holder}}}
   \depedge[edge start x offset=5pt]{1}{2}{\normalsize{\color{teal}{holder}}}
   \depedge{10}{12}{\normalsize{\color{red}{target}}}
   \depedge{10}{11}{\normalsize{\color{magenta}{exp:neg}}}
\end{dependency}
\caption{}
\end{subfigure}

\begin{subfigure}[head-first]{0.9\textwidth}
\centering
\smaller
\begin{dependency}[theme = simple]
   \begin{deptext}[column sep=1em]
     Some \& others \& give \& the \& new \& UMUC \& 5 \& stars \& - \& don't \& believe \& them.\\
   \end{deptext}
   \deproot{8}{\normalsize{\color{magenta}{exp:pos}}}
   \deproot{11}{\normalsize{\color{magenta}{exp:neg}}}
   \depedge[]{6}{4}{\normalsize{\color{red}{target}}}
   \depedge[]{6}{5}{\normalsize{\color{red}{target}}}
   \depedge[]{8}{6}{\normalsize{\color{red}{target}}}
   \depedge{8}{7}{\normalsize{\color{magenta}{exp:pos}}}
   \depedge[edge start x offset=5pt]{2}{1}{\normalsize{\color{teal}{holder}}}
   \depedge[edge start x offset=5pt]{8}{2}{\normalsize{\color{teal}{holder}}}
   \depedge{11}{12}{\normalsize{\color{red}{target}}}
   \depedge{11}{10}{\normalsize{\color{magenta}{exp:neg}}}
\end{dependency}
\caption{}
\end{subfigure}
\caption{Two parsing graph proposals to encode the sentiment graph: \textbf{(a) head-first}, where the first token of any span is the head, and \textbf{(b) head-final}, where the final token is the head. }
\label{fig:localgraphs}
\end{figure*}

\section{Modeling}

This section describes how we define and encode sentiment graphs, detail the neural dependency graph models, as well as two state-of-the-art baselines for end-to-end sentiment analysis (target and expression extraction, plus polarity classification).

\subsection{Graph Representations}
Structured sentiment graphs as in Figure~\ref{deptree} are directed graphs, that are made up of a set of labeled nodes and a set of unlabeled edges connecting pairs of nodes.
Nodes in the structured sentiment graphs can span over multiple tokens and may have multiple incoming edges.
The resulting graphs can have multiple entry points (\textit{roots}), are not necessarily connected, and not every token is a node in the graph.
The sentence's sentiment expressions correspond to the roots of the graphs, connecting explicitly to their respective holders and targets.
In order to apply the algorithm of \citet{dozat-manning-2018-simpler}, we simplify these structures into bi-lexical dependency graphs visualized in Figure~\ref{fig:localgraphs}.
Here, nodes correspond one-to-one to the tokens of the sequence and follow the same linear order.
The edges are drawn as arcs in the half-plane above the sentence, connecting \textit{heads} to \textit{dependents}.
Similarly to the source structures, the graphs can have multiple roots and nodes can have multiple or no incoming arcs.
For some rare instances of structured sentiment graphs, the reduction to dependency graphs is lossy, as they do not allow multiple arcs to share the same head and dependent.
This results in a slight mismatch of the learned and aimed-for representations.

The choice of how to encode the sentiment graphs as parsing graphs opens for several alternate representations depending on the choice of head/dependent status of individual tokens in the target/holder/expression spans of the sentiment graph.
We here propose two simple parsing graph representations: head-first and head-final, which are shown in Figure \ref{fig:localgraphs}.
For \textbf{head-first}, we set the first token of the sentiment expression as a root node, and similarly set the first token in each holder and token span as the head of the span with all other tokens within that span as dependents. The labels simply denote the type of relation (target/holder) and for sentiment expressions, additionally encode the polarity.
\textbf{Head-final} is similar, but instead sets the final token of spans as the heads, and the final token of the sentiment expression as the root node.

\begin{table}[]
{\smaller
    \centering
    \begin{tabular}{@{}lp{1.6cm}p{1.6cm}c@{}}
    \toprule
    Metric Name & Level & Strictness & $+$/$-$ \\
    \cmidrule(r){1-1}\cmidrule(lr){2-2}\cmidrule(lr){3-3}\cmidrule(l){4-4}
    \textbf{Holder \F} & Token-level & Partial & No\\
    \textbf{Target \F} & Token-level & Partial & No\\
    \textbf{Exp. \F} & Token-level & Partial & No\\
    \textbf{Targeted \F} & Token-level & Exact & Yes \\
    \textbf{U\F }& Graph arcs & Exact & No \\
    \textbf{L\F }& Graph arcs & Exact & Yes\\
    \textbf{NS\F} & Sentiment- graph & Exact graph, partial token & No \\
    \textbf{S\F} & Sentiment- graph & Exact graph, partial token & Yes \\
    \bottomrule
        \end{tabular}
    \caption{Metrics used to evaluate performance. Column $+$/$-$ indicates whether polarity is included or not. The main metrics are Targeted \F, which allows us to compare to methods that do not perform the full task, and S\F, which best represents the full task.}
    \label{tab:metrics}}
\end{table}

\subsection{Proposed model}

The neural graph parsing model used in this work is a reimplementation of the neural parser by \newcite{dozat-manning-2018-simpler} which was used by \newcite{kurtz-etal-2020-end} for negation resolution.
The parser learns to score each possible arc to then finally predict the output structure simply as a collection of all positively scored arcs.
The base of the network structure is a bidirectional LSTM (BiLSTM), that processes the input sentence both from left-to-right and right-to-left, to create contextualized representations $ c_1,\ldots,c_n = \textrm{BiLSTM}(w_1,\ldots,w_n)$
where $w_i$ is the concatenation of a word embedding, POS tag embedding, lemma embedding, and character embedding created by a character-based LSTM for the $i$th token.
In our experiments, we further augment the token representations with pre-trained contextualized embeddings from multilingual BERT \cite{xu-etal-2019-bert}.
We use multilingual BERT as several languages did not have available monolingual BERT models at the time of the experiments (Catalan, Norwegian).

The contextualized embeddings are then processed by two feedforward neural networks (FNN), creating specialized representations for potential heads and dependents, $h_i = \textrm{FNN}_{head}(c_i)$ and $d_i = \textrm{FNN}_{dep}(c_i)$.
The scores for each possible arc-label combination are computed by a final bilinear transformation using the tensor $U$.
Its inner dimension corresponds to the number of sentiment graph labels plus a special \texttt{NONE} label, indicating the absence of an arc, which allows the model to predict arcs and labels jointly, $\textrm{score}(h_i,d_j) = h_i^\top U d_j $.

\subsection{Baselines}
We compare our proposed graph prediction approach with three state-of-the-art baselines\footnote{Despite having state-of-the-art results on \mpqa, we do not compare with \newcite{katiyar-cardie-2016-investigating} as they use different dataset splits, 10-fold cross-validation, and their code is not available.} for extracting targets and expressions and predicting the polarity: \textbf{IMN}\footnote{IMN code available at \url{https://github.com/ruidan/IMN-E2E-ABSA}.}, \textbf{RACL}\footnote{\url{https://github.com/NLPWM-WHU/RACL}.}, as well as \textbf{RACL-BERT}, which also incorporates contextualized embeddings. Instead of using BERT$_{Large}$, we use the cased BERT-multilingual-base in order to fairly compare with our own models. Note, however, that our model does not update the mBERT representations, putting it at a disadvantage to RACL-BERT. We also compare with previously reported extraction results from \newcite{barnes-etal-2018-multibooked} and \newcite{ovrelid-etal-2020-fine}.

\begin{table*}[h!]
\renewcommand{\impr}[1]{\underline{\textbf{#1}}}
    \centering
    \small
    \resizebox{.95\textwidth}{!}{
    \begin{tabular}{llllllllll}
    \toprule
  Dataset & Model & \multicolumn{3}{c}{Spans}  & \multicolumn{1}{c}{Targeted} & \multicolumn{2}{c}{Parsing Graph} & \multicolumn{2}{c}{Sent. Graph} \\
    \cmidrule(lr){1-1}\cmidrule(lr){2-2}\cmidrule(lr){3-5}\cmidrule(lr){6-6}\cmidrule(lr){7-8}\cmidrule(lr){9-10}
        & & Holder \F & Target \F & Exp. \F & \F & U\F & L\F & NS\F & S\F \\
      \cmidrule(lr){3-5}\cmidrule(lr){6-6}\cmidrule(lr){7-8}\cmidrule(lr){9-10}
      
      \multirow{6}{*}{\norec}
    & \newcite{ovrelid-etal-2020-fine} & 42.4 & 31.3 & 31.3 &  - &  - & - & - & - \\
    & IMN & - & 35.9 & 48.7 & 18.0 & - & - & - & - \\
    &  RACL & - & 45.6 & 55.4 & 20.1 & - & - & - & - \\
    & RACL-BERT & - & 47.2 & \impr{56.3} & 30.3 & - & - & - & -\\
     \cmidrule(lr){3-5}\cmidrule(lr){6-6}\cmidrule(lr){7-8}\cmidrule(lr){9-10}
    & Head-first    & 51.1  & 50.1  & 54.4  & 30.5     & 39.2  & 31.5  & 37.0  & 29.5  \\
    & Head-final    & \impr{60.4}$\ast$ & \impr{54.8}  & 55.5  & \impr{31.9}     & \impr{48.0}$\ast$  & \impr{37.7}$\ast$  & \impr{39.2}$\ast$  & \impr{31.2}$\ast$ \\

    \cmidrule(lr){2-10}
      \multirow{6}{*}{\basque}
      & \newcite{barnes-etal-2018-multibooked}$^{\dagger}$ & 54.0 & 57.0 & 54.0 &  - &  - & - & - & - \\
      & IMN & - & 48.2 & 65.2 & 39.5 & - & - & - & - \\
      & RACL & - & 55.4 & 70.7 & 48.2& - & - & - & - \\
      & RACL-BERT & - & 59.9 & 72.6 & 56.8 & - & - & - & -\\
      \cmidrule(lr){3-5}\cmidrule(lr){6-6}\cmidrule(lr){7-8}\cmidrule(lr){9-10}

    & Head-first  & 60.4  & \impr{64.0}  & \impr{73.9}  & \impr{57.8}     & \impr{64.6}  & \impr{60.0}  & \impr{58.0}  & \impr{54.7}  \\
    & Head-final  & \impr{60.5 } & \impr{64.0}  & 72.1  & 56.9     & 60.8  & 56.0  & \impr{58.0}  & \impr{54.7}  \\

     \cmidrule(lr){2-10}
      \multirow{6}{*}{\cat}
      & \newcite{barnes-etal-2018-multibooked}$^{\dagger}$ & 56.0 & 64.0 & 52.0 & - &  - & - & - & - \\
      & IMN & - & 56.3 & 60.9 & 32.5 & - & - & - & - \\
      & RACL & - & 65.4 & 67.6 & 49.1& - & - & - & - \\
      & RACL-BERT & - & 67.5 & 70.3 & 52.4 & - & - & - & - \\
      \cmidrule(lr){3-5}\cmidrule(lr){6-6}\cmidrule(lr){7-8}\cmidrule(lr){9-10}

    & Head-first  & \impr{43.0}  & \impr{72.5}  & \impr{71.1}$\ast$ & \impr{55.0}$\ast$     & \impr{66.8}$\ast$ & \impr{62.1}$\ast$ & \impr{62.0 } & \impr{56.8 } \\
    & Head-final  & 37.1  & 71.2  & 67.1  & 53.9     & 62.7  & 58.1  & 59.7  & 53.7  \\

      \cmidrule(lr){2-10}
      \multirow{6}{*}{\mpqa}
      & IMN & - &  24.3 & 29.6 & 1.2 & - & - & - & - \\
      & RACL & - & 32.6 & 37.8 & 11.8& - & - & - & - \\
      & RACL-BERT & - & 20.0 & 31.2 & 17.8 & - & - & - & - \\
      \cmidrule(lr){3-5}\cmidrule(lr){6-6}\cmidrule(lr){7-8}\cmidrule(lr){9-10}

    & Head-first  & 43.8  & \impr{51.0}  & \impr{48.1 } & \impr{33.5}$\ast$     & 40.0  & 36.9  & 24.5  & 17.4  \\
    & Head-final  & \impr{46.3 } & 49.5  & 46.0  & 18.6    & \impr{41.4 } & \impr{38.0 } & \impr{26.1 } & \impr{18.8 } \\

      \cmidrule(lr){2-10}
      \multirow{6}{*}{\dsu}
      & IMN & - & 33.0 & 27.4 & 17.9 & - & - & - & - \\
      & RACL & - & 39.3 & 40.2 & 22.8& - & - & - & - \\
      & RACL-BERT & - & \impr{44.6} & 38.2 & 27.3 & - & - & - & - \\
      \cmidrule(lr){3-5}\cmidrule(lr){6-6}\cmidrule(lr){7-8}\cmidrule(lr){9-10}

        & Head-first & 28.0   & 39.9  & 40.3  & 26.7     & 35.3  & 31.4  & 31.0  & 25.0  \\
    & Head-final & \impr{37.4 } & 42.1 & \impr{45.5}$\ast$ & \impr{29.6 }    & \impr{38.1 } & \impr{33.9 } & \impr{34.3}$\ast$ & \impr{26.5}  \\

    \bottomrule
    \end{tabular}
    }
    \caption{Experiments comparing our sentiment graph approaches (Head-first/Head-final) using mBERT with the sequence-labeling baselines (IMN, RACL, RACL-BERT). \impr{Underlined} numbers indicate the best result for the metric and dataset. $\ast$ indicates approach is significantly better than second best ($p<0.05$), as determined by a bootstrap with replacement test. $^{\dagger}$ indicates results that are not comparable, as they were calculated with 10-fold cross-validation.}
    \label{tab:main_results}
\end{table*}

\section{Evaluation}
    
As we are interested not only in extraction or classification, but rather in the full structured sentiment task, we propose metrics that capture the relations between all predicted elements, while enabling comparison with previous state-of-the-art models on different subtasks. The \underline{main metrics} we use to rank models are Targeted \F and Sentiment Graph \F. 

\paragraph{Token-level \F for Holders, Targets, and Expressions}
To easily compare our models to pipeline models, we evaluate how well these models are able to identify the elements of a sentiment graph with token-level \F.

\paragraph{Targeted \F} This is a common metric in targeted sentiment analysis (also referred to as \F-i \cite{he-etal-2019-interactive} or ABSA \F \cite{chen-qian-2020-relation}). A true positive requires the combination of exact extraction of the sentiment target, and the correct polarity.

\paragraph{Parsing graph metrics}
We additionally compute graph-level metrics to determine how well the models predict the unlabeled and labeled arcs of the parsing graphs: Unlabeled \F (\textbf{U\F}), Labeled \F (\textbf{L\F}).
These measure the amount of (in)correctly predicted arcs and labels, as the harmonic mean of precision and recall \cite{oepen2014semeval}.
These metrics inform us of the local properties of the graph, and do not overly penalize a model if a few edges of a graph are incorrect.

\paragraph{Sentiment graph metrics}

The two metrics that measure how well a model is able to capture the full sentiment graph (see Figure \ref{deptree}) are
Non-polar Sentiment Graph \F (\textbf{NS\F}) and Sentiment Graph \F (\textbf{S\F}).
For \textbf{NS\F}, each sentiment graph is a tuple of (holder, target, expression), while for \textbf{S\F} we include polarity (holder, target, expression, polarity). A true positive is defined as an exact match at graph-level, weighting the overlap in predicted and gold spans for each element, averaged across all three spans. 
For precision we weight the number of correctly predicted tokens divided by the total number of predicted tokens (for recall, we divide instead by the number of gold tokens). We allow for empty holders and targets.

\section{Experiments}

All sentiment graph models use token-level mBERT representations in addition to word2vec skip-gram embeddings openly available from the NLPL vector repository\footnote{Nordic Language Processing Laboratory vector repo.: \url{http://vectors.nlpl.eu/repository/}. We used 300-dimensional embeddings trained on English Wikipedia and Gigaword for English (model id 18 in the repo.), and 100-dimensional embeddings trained on the 2017 CoNLL corpora for all others; Basque (id 32),  Catalan (id 34), and Norwegian Bokmål (id 58).}  \cite{fares-etal-2017-word}. We train all models for 100 epochs and keep the model that performs best regarding L\F on the dev set (Targeted \F for the baselines). We use default hyperparameters from \newcite{kurtz-etal-2020-end} (see Appendix) and run all of our models five times with different random seeds and report the mean (standard deviation shown as well in Table \ref{tab:all_others} in the Appendix). We calculate statistical difference between the best and second best models through a bootstrap with replacement test \cite{berg-kirkpatrick2012empirical}. As there are 5 runs, we require that 3 of 5 be statistically significant at $p<0.05$. Table~\ref{tab:main_results} shows the results for all datasets. 

 On \norec, the baselines IMN, RACL, and RACL-BERT perform well at extracting targets (35.9, 45.6, and 47.2 \F, respectively) and expressions (48.7/55.4/56.3),
 but struggle with the full targeted sentiment task (18.0/20.1/30.3). The graph-based models extract targets better (50.1/54.8) and have comparable scores for expressions (54.4/55.5). The holder extraction scores have a similar range (51.1/60.4). These patterns hold throughout the other datasets, where the proposed graph models nearly always perform best on extracting spans, although RACL-BERT achieves the best score on extracting targets on \dsu (44.6 vs. 42.1). The graph models also outperform the strongest baseline (RACL-BERT) on \emph{targeted sentiment} on all 5 datasets, although this difference is often not statistically significant (\norec Head-first, \basque Head-final) and RACL-BERT is better than Head-first on \dsu.
 
 Regarding the Graph metrics, the results depend highly on the dataset, with U\F and L\F ranging from 35.3/31.4 (\dsu Head-first) to 66.8/62.1 (\cat Head-first). Sentiment Graph metrics NS\F and S\F have a similar, though slightly lower range (24.5/17.7 -- 62.0/56.8). The graph and sentiment graph metrics do not correlate perfectly, however, as U\F and L\F on \mpqa are relatively good (40.0/36.9 and 41.4/38.0 for Head-first and Head-final, respectively), but the NS\F and S\F are poor (24.5/17.4 and 26.1/18.8).
 
 On average IMN is the weakest baseline, followed by RACL and then RACL-BERT. The main improvement that RACL-BERT gives over RACL on these datasets is seen in the Targeted metric, \ie, the contextualized representations improve the polarity classification more than the extraction task. 
 The proposed graph-based models are consistently the best models across the metrics and datasets.

Regarding graph representations, the differences between Head-first and Head-final are generally quite small. Head-first performs better on \cat and slightly better on \basque, while for the others (\norec, \mpqa, and \dsu) Head-final is better. This suggests that the main benefit is the joint prediction of all spans and relationships, and that the specific graph representation matters less.

\section{Analysis}

\begin{table}[]
 {\smaller
    \centering
    \begin{tabular}{@{}lrrrr@{}}
    \toprule
    & \# & H.first & H.final & RACL \\
    \cmidrule(lr){2-2}\cmidrule(lr){3-3}\cmidrule(lr){4-4}\cmidrule(lr){5-5}
        \norec & 147 & 63.3 &67.8 & 65.6 \\
        \basque & 45 & 68.9 &65.9 & 29.2\\
        \cat & 74 & 72.2 &73.7 & 28.2\\
        \mpqa & 40 & 55.4 &58.5 & 28.8\\
        \dsu & 10 & 56.9 &43.1 & 31.4\\
    \bottomrule
    \end{tabular}
    \caption{Number of sentences with multiple targets (\#) and Macro \F on the target extraction task for Head-final and RACL. Head-final is consistently better than RACL on extracting multiple targets. }
    \label{tab:mult_targs}}
\end{table}

\begin{table*}[]
    \centering
    \small
    \begin{tabular}{@{}lcccccccc@{}}
    \toprule
   &  \multicolumn{3}{c}{Spans}  & \multicolumn{1}{c}{Targeted} & \multicolumn{2}{c}{Graph} & \multicolumn{2}{c}{Sent. Graph} \\
    \cmidrule(lr){2-4}\cmidrule(lr){5-5}\cmidrule(lr){6-7}\cmidrule(lr){8-9}
        &  Holder \F & Target \F & Exp. \F & \F & U\F & L\F & NS\F & S\F \\
      \cmidrule(lr){2-4}\cmidrule(lr){5-5}\cmidrule(lr){6-7}\cmidrule(lr){8-9}
\norec & 1.2 & 5.0 & 3.4 & 4.2 & 2.8 & 2.7 & 4.6 & 4.0\\
\basque & 2.9 & 0.6 & 0.8 & 1.1 & 1.0 & 1.4 & 1.2 & 1.4\\
\cat & 0.4 & 1.6 & 1.6 & 2.1 & 2.0 & 1.8 & 3.3 & 2.8\\
\mpqa & 8.2 & 8.8 & 5.2 & 7.2 & 6.6 & 7.3 & 5.4 & 5.1\\
\dsu & 7.9 & 1.2 & 4.3 & 6.4 & 3.9 & 5.7 & 3.6 & 6.0\\
    \bottomrule
    \end{tabular}
    \caption{Average gains in percentage points by including mBERT representations.}
    \label{tab:adding_bert}
\end{table*}

In this section we perform a deeper analysis of the models in order to answer the research questions.

\subsection{Do syntactically informed sentiment graphs improve results?} 
Our two baseline graph representations, Head-first and Head-final, are crude approximations of linguistic structure. In syntactic and semantic dependency graphs, heads are often neither the first or last word, but rather the most salient word according to various linguistic criteria. 
First, we enrich the dependency labels to distinguish edges that are internal to a holder/target/expression span from those that are external and perform experiments by adding an `in label' to non-head nodes within the graph, which we call \textbf{+inlabel}.
     We further inform the head selection of the parsing graphs with syntactic information in  the \textbf{Dep. edges} parsing graphs, where we compute the dependency graph for each sentence\footnote{We use SpaCy \cite{spacy} for English, Stanza \cite{qi-etal-2020-stanza} for Basque and Catalan and UDPipe \cite{udpipe:2017} for Norwegian.} and set the head of each span to be the node that has an outgoing edge in the corresponding syntactic graph. As there can be more than one such edge, we default to the first. A manual inspection showed that this approach sometimes set unlikely dependency label types as heads, \eg, \texttt{punct}, \texttt{obl}. Therefore, we suggest a final approach, \textbf{Dep. labels}, which filters out these unlikely heads. The full results are shown in Table~\ref{tab:all_others} in the Appendix. 
\begin{figure}
    \centering
    \includegraphics[width=.45\textwidth]{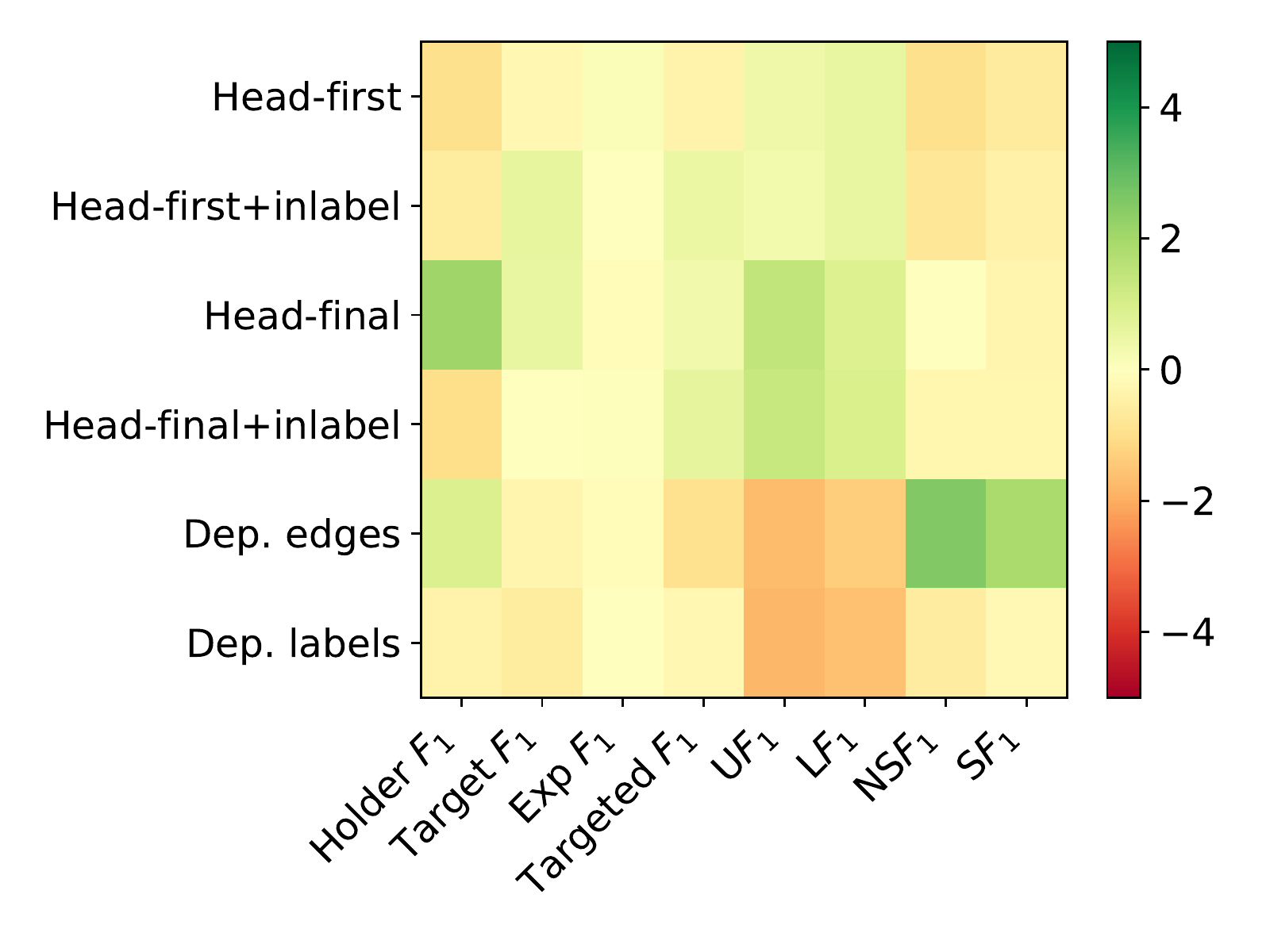}
    \caption{Average benefit of each graph annotation scheme (y-axis) on the evaluation metrics (x-axis) in percentage points. The results are averaged across datasets.}
    \label{fig:aggregate}
\end{figure}
The implementation of the graph structure has a large effect on all metrics, although the specific results depend on the dataset. We plot the average effect of each implementation across all datasets in Figure \ref{fig:aggregate}, as well as each individual dataset (Figures \ref{fig:aggregate_norec}--\ref{fig:aggregate_dsunis} in the Appendix). 
\textbf{+inlabel} tends to improve results on the non-English datasets, consistently increasing target and expression extraction 
and targeted sentiment.
It also generally improves the graph scores U\F and L\F on the non-English datasets. 

Dep. edges has the strongest positive effect on the NS\F and S\F (an avg. 2.52 and  2.22 percentage point (pp) over Head-final, respectively). However, this average is pulled down by poorer performance on the English datasets. Removing these two, the average benefit is 5.2 and 4.2 for NS\F and S\F, respectively. On span extraction and targeted sentiment, however, Dep. edges leads to poorer scores overall. Dep. labels does not lead to any consistent improvements.
These results indicate that incorporating syntactic dependency information is particularly helpful for the full structured sentiment task, but that these benefits do not always show at a more local level, \ie, span extraction.

\subsection{Do graph models perform better on sentences with multiple targets?}

We hypothesize that predicting the full sentiment graph may have a larger effect on sentences with multiple targets. Therefore, we create a subset of the test data containing sentences with multiple targets and reevaluate Head-first, Head-final, and RACL-BERT on the target extraction task. Table~\ref{tab:mult_targs} shows the number of sentences with multiple targets and the Target span extraction score for each model. On this subset, Head-first and Head-final outperform RACL-BERT on 9 of 10 experiments, confirming the hypothesis that the graph models improve on examples with multiple targets.

\subsection{How much does mBERT contribute?}

We also perform experiments without mBERT (shown in Table~\ref{tab:non-contextualized} in the Appendix) and show the average gains (over all 6 graph setups) of including it in Table~\ref{tab:adding_bert}. 
Adding the mBERT features leads to average improvements in all experiments: for extracting spans an average gain of 4.1 pp for holders, 3.4 for targets, and 3.1 for expressions. For targeted sentiment there is a larger gain of 4.2 pp, while for the parsing graph metrics U\F and l\F the gains are more limited (3.3 pp/ 3.8 pp) and similarly for NS\F and S\F (3.6 pp/ 3.9 pp). 
The gains are largest for the English datasets (\mpqa, \dsu) followed by \norec, and finally \cat and \basque. This corroborates the bias towards English and similar languages that has been found in multilingual language models \cite{artetxe-etal-2020-cross,conneau-etal-2020-unsupervised} and motivates the need for language-specific contextualized embeddings.

\subsection{Analysis of polarity predictions}

In this section we zoom in on polarity, in order to quantify how well models perform at predicting only polarity. As the polarity annotations are bound to the expressions, we consider true positives to be any expression that overlaps the gold expression and has the same polarity. Table \ref{tab:polarity} shows that the polarity predictions are best on \basque and \cat, followed by \norec and \dsu, and finally \mpqa. This is likely due to the number of domains and characteristics of the data. \norec contains many domains and has longer expressions, while \mpqa contains many highly ambiguous polar expressions, \eg, `said', `asked', which have different polarity depending on the context.

\begin{table}[]
    \centering
\begin{tabular}{llll}
\toprule
  \norec  & 57.0 \scriptsize{(1.5)} \\
  \basque &  75.7 \scriptsize{(0.8)} \\
 \cat & 71.7 \scriptsize{(2.4)}  \\ 
\mpqa & 38.5 \scriptsize{(1.4)} \\ 
\dsu & 44.5 \scriptsize{(2.4)} \\
\bottomrule
\end{tabular}
    \caption{Polarity \F scores (unweighted and weighted) of models augmented with mBERT on the head-final setup. We report average and standard deviation over 5 runs. }
    \label{tab:polarity}
\end{table}

\section{Conclusion}
In this paper, we have proposed a dependency graph parsing approach to structured sentiment analysis and shown that these models outperform state-of-the-art sequence labeling models on five benchmark datasets. 

Using parse trees as input has shown promise for sentiment analysis in the past, either to guide a tree-based algorithm \cite{socher-etal-2013-recursive,tai-etal-2015-improved} or to create features for sentiment models \cite{nakagawa-etal-2010-dependency,almeida-etal-2015-aligning}. However, to the authors' knowledge, this is the first attempt to directly predict dependency-based sentiment graphs. 

In the future, we would like to better exploit the similarities between dependency parsing and sentiment graph parsing, either by augmenting the token-level representations with contextualized vectors from their heads in a dependency tree \cite{kurtz-etal-2020-end} or by multi-task learning to dependency parse.
We would also like to explore different graph parsing approaches, \eg, PERIN \cite{samuel-straka-2020-ufal}.

\section*{Acknowledgements}
This work has been carried out as part of the SANT project (Sentiment Analysis for Norwegian Text), funded by the Research Council of Norway (grant number 270908). 

The computations were performed on resources provided by 
UNINETT Sigma2 - the National Infrastructure for High Performance Computing and Data Storage in Norway.

\bibliographystyle{acl_natbib}
\bibliography{anthology,emnlp2020,robins_bib}

\clearpage
\appendix
\section{Appendix}

\begin{table*}[]
    \centering
    \small
    \resizebox{\textwidth}{!}{
    \begin{tabular}{llrrrrrrrr}
    \toprule
   & & \multicolumn{3}{c}{Spans}  & \multicolumn{1}{c}{Targeted} & \multicolumn{2}{c}{Graph} & \multicolumn{2}{c}{Sent. Graph} \\
    \cmidrule(lr){3-5}\cmidrule(lr){6-6}\cmidrule(lr){7-8}\cmidrule(lr){9-10}
        & & Holder \F & Target \F & Exp. \F & \F & U\F & L\F & NS\F & S\F \\
      \cmidrule(lr){3-5}\cmidrule(lr){6-6}\cmidrule(lr){7-8}\cmidrule(lr){9-10}
      
      \multirow{8}{*}{\rotatebox{90}{\norec}}
    & IMN & - & 35.9 & 48.7 & 18.0 & - & - & - & - \\
    &  RACL & - & 45.6 & \impr{55.4} & 20.1 & - & - & - & - \\
    \cmidrule(lr){2-10}
    & Head-first    & 48.4 \scriptsize{(2.2)} & 47.1 \scriptsize{(1.6)} & 52.0 \scriptsize{(1.6)} & \impr{33.0 \scriptsize{(1.4)}}    & 37.6 \scriptsize{(0.5)} & 29.8 \scriptsize{(0.4)} & 32.9 \scriptsize{(1.6)} & 26.1 \scriptsize{(1.5)} \\
    & +inlabel      & 50.4 \scriptsize{(4.0)} & 47.6 \scriptsize{(2.5)} & 51.0 \scriptsize{(1.3)} & 27.3 \scriptsize{(1.1)}    & 36.9 \scriptsize{(0.5)} & 29.4 \scriptsize{(0.8)} & 32.9 \scriptsize{(1.1)} & 25.8 \scriptsize{(0.5)} \\
    & Head-final    & 57.0 \scriptsize{(3.3)} & \impr{49.4 \scriptsize{(0.9)}} & 52.1 \scriptsize{(1.8)} & 26.0 \scriptsize{(0.6)}    & \impr{45.1 \scriptsize{(1.2)}} & \impr{35.2 \scriptsize{(1.1)}} & 34.4 \scriptsize{(0.7)} & 27.2 \scriptsize{(0.9)} \\
    & +inlabel      & \impr{57.9 \scriptsize{(1.7)}} & 50.1 \scriptsize{(1.3)} & 52.6 \scriptsize{(0.4)} & 29.6 \scriptsize{(0.6)}    & 45.0 \scriptsize{(1.0)} & \impr{35.2 \scriptsize{(0.5)}} & 35.1 \scriptsize{(1.6)} & 27.0 \scriptsize{(1.3)} \\
    & Dep. edges    & 54.4 \scriptsize{(3.9)} & 49.0 \scriptsize{(2.5)} & 51.4 \scriptsize{(1.7)} & 26.7 \scriptsize{(3.1)}    & 39.3 \scriptsize{(1.1)} & 31.5 \scriptsize{(1.3)} & \impr{47.2 \scriptsize{(0.9)}} & \impr{36.0 \scriptsize{(1.1)}} \\
    & Dep. labels   & 51.6 \scriptsize{(2.6)} & 46.5 \scriptsize{(3.0)} & 50.7 \scriptsize{(2.7)} & 26.7 \scriptsize{(1.9)}    & 36.7 \scriptsize{(1.1)} & 28.3 \scriptsize{(0.8)} & 33.4 \scriptsize{(1.8)} & 25.4 \scriptsize{(1.8)} \\

    \midrule
      \multirow{8}{*}{\rotatebox{90}{\basque}}
      & IMN & - & 48.2 & 65.2 & 39.5 & - & - & - & - \\
      & RACL & - & 55.4 & 70.7 & 48.2\\
      \cmidrule(lr){2-10}
    & Head-first    & \impr{60.8 \scriptsize{(3.8)}}  & 64.1 \scriptsize{(1.4)}  & 72.2 \scriptsize{(0.7)}  & 53.9 \scriptsize{(1.8)}     & \impr{62.9 \scriptsize{(0.6)}}  & \impr{58.2 \scriptsize{(0.3)}}  & 58.5 \scriptsize{(2.3)}  & 54.7 \scriptsize{(2.6)}  \\
    & +inlabel      & 59.8 \scriptsize{(1.6)}  & 64.3 \scriptsize{(0.9)}  & 71.9 \scriptsize{(0.8)}  & \impr{57.9 \scriptsize{(1.9)}}     & 62.6 \scriptsize{(0.6)}  & 57.5 \scriptsize{(1.1)}  & 57.3 \scriptsize{(1.5)}  & 53.6 \scriptsize{(1.3)}  \\
    & Head-final    & 57.0 \scriptsize{(2.0)}  & \impr{66.0 \scriptsize{(1.6)}}  & 72.2 \scriptsize{(0.6)}  & 55.5 \scriptsize{(1.7)}     & 60.2 \scriptsize{(0.8)}  & 55.5 \scriptsize{(0.9)}  & \impr{59.6 \scriptsize{(0.8)}}  & \impr{56.3 \scriptsize{(1.0)}}  \\
    & +inlabel      & 53.7 \scriptsize{(1.2)}  & 64.0 \scriptsize{(2.4)}  & \impr{72.9 \scriptsize{(0.6)}}  & 54.9 \scriptsize{(2.0)}     & 60.1 \scriptsize{(1.5)}  & 54.9 \scriptsize{(1.7)}  & 57.1 \scriptsize{(3.2)}  & 53.5 \scriptsize{(3.3)}  \\
    & Dep. edges    & 53.1 \scriptsize{(1.9)}  & 63.8 \scriptsize{(1.7)}  & 71.0 \scriptsize{(1.1)}  & 53.7 \scriptsize{(1.7)}     & 59.0 \scriptsize{(1.3)}  & 54.5 \scriptsize{(1.6)}  & 59.0 \scriptsize{(1.6)}  & 55.6 \scriptsize{(1.8)}  \\
    & Dep. labels   & 52.0 \scriptsize{(3.8)}  & 63.0 \scriptsize{(1.1)}  & 71.3 \scriptsize{(1.9)}  & 54.0 \scriptsize{(1.0)}     & 59.5 \scriptsize{(1.1)}  & 54.9 \scriptsize{(1.1)}  & 58.6 \scriptsize{(2.8)}  & 54.6 \scriptsize{(2.4)}  \\

      \midrule
      \multirow{8}{*}{\rotatebox{90}{\cat}}
      & IMN & - & 56.3 & 60.9 & 32.5 & - & - & - & - \\
      & RACL & - & 65.4 & 67.6 & 53.1\\
      \cmidrule(lr){2-10}

    & Head-first    & 41.9 \scriptsize{(2.8)} & 69.8 \scriptsize{(1.7)} & 68.9 \scriptsize{(1.4)} & \impr{57.3 \scriptsize{(2.0)}}    & 64.2 \scriptsize{(0.7)} & \impr{59.9 \scriptsize{(0.8)}} & \impr{58.2 \scriptsize{(2.3)}} & 53.3 \scriptsize{(2.2)} \\
    & +inlabel      & 42.4 \scriptsize{(2.6)} & \impr{70.9 \scriptsize{(0.8)}} & \impr{69.9 \scriptsize{(0.9)}} & 50.9 \scriptsize{(1.3)}    & \impr{64.4 \scriptsize{(0.9)}} & 59.6 \scriptsize{(0.7)} & 55.7 \scriptsize{(2.0)} & 50.7 \scriptsize{(2.1)} \\
    & Head-final    & 40.4 \scriptsize{(2.5)} & 69.9 \scriptsize{(1.5)} & 66.8 \scriptsize{(0.8)} & 50.8 \scriptsize{(2.6)}    & 60.9 \scriptsize{(0.6)} & 57.1 \scriptsize{(0.8)} & 57.7 \scriptsize{(1.0)} & 53.3 \scriptsize{(1.3)} \\
    & +inlabel      & 36.4 \scriptsize{(2.2)} & 69.1 \scriptsize{(1.1)} & 65.4 \scriptsize{(0.6)} & 52.9 \scriptsize{(0.9)}    & 60.6 \scriptsize{(0.9)} & 57.0 \scriptsize{(0.6)} & 58.0 \scriptsize{(1.9)} & \impr{53.5 \scriptsize{(2.0)}} \\
    & Dep. edges    & 42.6 \scriptsize{(6.1)} & 69.1 \scriptsize{(0.5)} & 67.3 \scriptsize{(0.6)} & 50.6 \scriptsize{(1.3)}    & 59.3 \scriptsize{(0.7)} & 55.7 \scriptsize{(1.1)} & 57.5 \scriptsize{(2.1)} & 52.8 \scriptsize{(1.8)} \\
    & Dep. labels   & \impr{43.8 \scriptsize{(3.4)}} & 70.3 \scriptsize{(0.8)} & 67.2 \scriptsize{(1.1)} & 50.6 \scriptsize{(1.8)}    & 61.0 \scriptsize{(0.5)} & 57.1 \scriptsize{(0.8)} & 57.8 \scriptsize{(1.4)} & 52.7 \scriptsize{(1.9)} \\

      \midrule
      \multirow{8}{*}{\rotatebox{90}{\mpqa}}
      & IMN & - &  24.3 & 29.6 & 1.2 & - & - & - & - \\
      & RACL & - & 32.6 & 37.8 & 11.8\\
      \cmidrule(lr){2-10}

    & Head-first    & 35.2 \scriptsize{(1.1)} & 40.5 \scriptsize{(1.8)} & 41.7 \scriptsize{(1.7)} & \impr{22.6 \scriptsize{(3.1)}}    & 32.2 \scriptsize{(1.3)} & 28.2 \scriptsize{(1.4)} & 19.4 \scriptsize{(1.5)} & 12.4 \scriptsize{(1.7)} \\
    & +inlabel      & 35.6 \scriptsize{(1.4)} & 41.6 \scriptsize{(1.1)} & \impr{42.4 \scriptsize{(2.3)}} & 14.0 \scriptsize{(0.9)}    & 32.9 \scriptsize{(0.8)} & 28.9 \scriptsize{(0.9)} & \impr{20.4 \scriptsize{(1.0)}} & \impr{13.2 \scriptsize{(1.2)}} \\
    & Head-final    & \impr{37.1 \scriptsize{(1.3)}} & 42.1 \scriptsize{(1.0)} & 41.9 \scriptsize{(0.8)} & 13.3 \scriptsize{(1.9)}    & \impr{35.7 \scriptsize{(0.8)}} & \impr{31.7 \scriptsize{(0.5)}} & 18.7 \scriptsize{(0.7)} & 12.5 \scriptsize{(1.6)} \\
    & +inlabel      & 37.0 \scriptsize{(0.5)} & \impr{42.3 \scriptsize{(1.2)}} & 41.6 \scriptsize{(1.6)} & 15.2 \scriptsize{(2.5)}    & 35.5 \scriptsize{(0.7)} & 31.7 \scriptsize{(0.6)} & 19.6 \scriptsize{(0.6)} & 12.6 \scriptsize{(0.8)} \\
    & Dep. edges    & 35.4 \scriptsize{(1.5)} & 39.1 \scriptsize{(2.0)} & 41.6 \scriptsize{(1.1)} & 12.3 \scriptsize{(0.8)}    & 28.9 \scriptsize{(1.7)} & 24.8 \scriptsize{(1.5)} & 19.0 \scriptsize{(0.9)} & 11.9 \scriptsize{(1.1)} \\
    & Dep. labels   & 36.7 \scriptsize{(0.6)} & 39.2 \scriptsize{(2.3)} & 40.4 \scriptsize{(2.1)} & 12.6 \scriptsize{(1.2)}    & 28.9 \scriptsize{(1.6)} & 24.5 \scriptsize{(1.1)} & 18.9 \scriptsize{(0.8)} & 11.6 \scriptsize{(1.0)} \\

      \midrule
      \multirow{8}{*}{\rotatebox{90}{\dsu}}
      & IMN & - & 33.0 & 27.4 & 17.9 & - & - & - & - \\
      & RACL & - & 39.3 & 40.2 & 22.8\\
      \cmidrule(lr){2-10}

    & Head-first    & 25.6 \scriptsize{(5.1)}  & 36.8 \scriptsize{(3.5)} & 39.0 \scriptsize{(1.5)} & 23.4 \scriptsize{(1.8)}    & 32.9 \scriptsize{(1.7)} & 25.9 \scriptsize{(1.7)} & 27.8 \scriptsize{(1.4)} & 18.8 \scriptsize{(2.0)} \\
    & +inlabel      & 22.9 \scriptsize{(5.7)}   & 38.6 \scriptsize{(3.9)} & 38.6 \scriptsize{(2.8)} & 18.3 \scriptsize{(2.7)}    & \impr{33.7 \scriptsize{(2.8)}} & 26.4 \scriptsize{(1.6)} & 25.9 \scriptsize{(3.4)} & 15.2 \scriptsize{(2.2)} \\
    & Head-final    & 29.2 \scriptsize{(8.4)}  & 38.1 \scriptsize{(2.0)} & \impr{39.5 \scriptsize{(2.4)}} & 21.8 \scriptsize{(1.1)}    & 31.1 \scriptsize{(2.1)} & 26.0 \scriptsize{(1.0)} & 29.1 \scriptsize{(3.3)} & 20.4 \scriptsize{(2.0)} \\
    & +inlabel      & 30.2 \scriptsize{(8.9)}  & 38.2 \scriptsize{(2.9)} & 38.8 \scriptsize{(2.0)} & \impr{24.4 \scriptsize{(2.7)}}    & 32.4 \scriptsize{(2.1)} & \impr{28.4 \scriptsize{(2.1)}} & 28.1 \scriptsize{(2.9)} & \impr{22.4 \scriptsize{(3.4)}} \\
    & Dep. edges    & \impr{33.9 \scriptsize{(5.4)}}  & 39.2 \scriptsize{(2.9)} & 39.3 \scriptsize{(3.4)} & 22.2 \scriptsize{(2.7)}    & 32.4 \scriptsize{(2.1)} & 26.8 \scriptsize{(2.0)} & \impr{29.3 \scriptsize{(1.5)}} & 19.8 \scriptsize{(1.2)} \\
    & Dep. labels   & 21.3 \scriptsize{(18.1)} & \impr{40.0 \scriptsize{(1.5)}} & 38.4 \scriptsize{(2.4)} & 21.4 \scriptsize{(4.3)}    & 32.1 \scriptsize{(1.5)} & 27.2 \scriptsize{(1.5)} & 28.2 \scriptsize{(1.1)} & 20.5 \scriptsize{(3.1)} \\

    \bottomrule
    \end{tabular}
    }
    \caption{Experiments without contextualized embeddings.}
    \label{tab:non-contextualized}
\end{table*}

\begin{table*}[]
    \centering
    \small
    \resizebox{1\textwidth}{!}{
    \begin{tabular}{llrrrrrrrr}
    \toprule
   & & \multicolumn{3}{c}{Spans}  & \multicolumn{1}{c}{Targeted} & \multicolumn{2}{c}{Graph} & \multicolumn{2}{c}{Sent. Graph} \\
    \cmidrule(lr){3-5}\cmidrule(lr){6-6}\cmidrule(lr){7-8}\cmidrule(lr){9-10}
        & & Holder \F & Target \F & Exp. \F & \F. & U\F & L\F & NS\F & S\F \\
      \cmidrule(lr){3-5}\cmidrule(lr){6-6}\cmidrule(lr){7-8}\cmidrule(lr){9-10}
      
      \multirow{8}{*}{\rotatebox{90}{\norec}}

    & RACL-BERT & - & 47.2 & 56.3 & 30.3 & - & - & - & -\\
    \cmidrule(lr){2-10}
    & Head-first    & 51.1 \scriptsize{(3.2)} & 50.1 \scriptsize{(3.4)} & 54.4 \scriptsize{(1.6)} & 30.5 \scriptsize{(2.3)}    & 39.2 \scriptsize{(0.5)} & 31.5 \scriptsize{(0.5)} & 37.0 \scriptsize{(2.6)} & 29.5 \scriptsize{(2.4)} \\
    & +inlabel      & 51.6 \scriptsize{(2.8)} & 52.7 \scriptsize{(0.7)} & 54.6 \scriptsize{(1.4)} & 32.2 \scriptsize{(1.4)}    & 39.6 \scriptsize{(0.8)} & 32.0 \scriptsize{(0.7)} & 37.6 \scriptsize{(1.2)} & 29.5 \scriptsize{(1.2)} \\
    & Head-final    & \impr{60.4 \scriptsize{(1.2)}} & 54.8 \scriptsize{(1.6)} & 55.5 \scriptsize{(1.5)} & 31.9 \scriptsize{(1.3)}    & 48.0 \scriptsize{(1.3)} & 37.7 \scriptsize{(1.4)} & 39.2 \scriptsize{(1.7)} & 31.2 \scriptsize{(1.6)} \\
    & +inlabel      & 57.1 \scriptsize{(3.0)} & \impr{55.2 \scriptsize{(1.0)}} & \impr{56.3 \scriptsize{(1.3)}} & \impr{34.8 \scriptsize{(1.0)}}    & \impr{48.7 \scriptsize{(1.2)}} & \impr{38.3 \scriptsize{(1.0)}} & 40.5 \scriptsize{(1.1)} & 31.7 \scriptsize{(1.1)} \\
    & Dep. edges    & 54.0 \scriptsize{(3.4)} & 53.6 \scriptsize{(1.5)} & 55.0 \scriptsize{(0.9)} & 32.7 \scriptsize{(1.6)}    & 41.5 \scriptsize{(0.7)} & 33.8 \scriptsize{(0.4)} & \impr{50.9 \scriptsize{(0.3)}} & \impr{39.4 \scriptsize{(0.4)}} \\
    & Dep. labels   & 52.7 \scriptsize{(5.6)} & 53.6 \scriptsize{(0.3)} & 54.4 \scriptsize{(1.5)} & 32.7 \scriptsize{(1.6)}    & 40.7 \scriptsize{(0.8)} & 32.2 \scriptsize{(0.5)} & 38.2 \scriptsize{(1.4)} & 30.0 \scriptsize{(1.2)} \\
    
    \midrule
      \multirow{8}{*}{\rotatebox{90}{\basque}}

      & RACL-BERT & - & 59.9 & 72.6 & 56.8 & - & - & - & -\\
      \cmidrule(lr){2-10}

    & Head-first  & 60.4 \scriptsize{(2.2)} & 64.0 \scriptsize{(2.4)} & 73.9 \scriptsize{(1.0)} & 57.8 \scriptsize{(2.4)}    & 64.6 \scriptsize{(1.0)} & 60.0 \scriptsize{(1.6)} & 58.0 \scriptsize{(1.1)} & 54.7 \scriptsize{(1.6)} \\
    & +inlabel    & 59.6 \scriptsize{(1.9)} & \impr{65.9 \scriptsize{(0.9)}} & \impr{74.2 \scriptsize{(0.7)}} & \impr{59.2 \scriptsize{(0.9)}}    & \impr{64.7 \scriptsize{(0.7)}} & \impr{60.3 \scriptsize{(1.1)}} & 59.8 \scriptsize{(1.1)} & 56.1 \scriptsize{(1.6)} \\
    & Head-final  & \impr{60.5 \scriptsize{(2.2)}} & 64.0 \scriptsize{(2.3)} & 72.1 \scriptsize{(1.2)} & 56.9 \scriptsize{(1.7)}    & 60.8 \scriptsize{(0.8)} & 56.0 \scriptsize{(1.1)} & 58.0 \scriptsize{(2.1)} & 54.7 \scriptsize{(1.8)} \\
    & +inlabel    & 58.1 \scriptsize{(2.4)} & 64.7 \scriptsize{(1.1)} & 72.0 \scriptsize{(0.7)} & 58.5 \scriptsize{(1.4)}    & 60.6 \scriptsize{(1.1)} & 56.6 \scriptsize{(0.7)} & 59.8 \scriptsize{(1.6)} & 56.9 \scriptsize{(1.8)} \\
    & Dep. edges  & 58.8 \scriptsize{(4.2)} & 64.8 \scriptsize{(1.4)} & 71.2 \scriptsize{(0.8)} & 54.0 \scriptsize{(1.8)}    & 59.9 \scriptsize{(0.4)} & 55.5 \scriptsize{(0.7)} & \impr{60.9 \scriptsize{(1.6)}} & \impr{57.4 \scriptsize{(1.6)}} \\
    & Dep. labels & 56.3 \scriptsize{(2.1)} & 65.4 \scriptsize{(0.9)} & 72.9 \scriptsize{(1.1)} & 54.9 \scriptsize{(0.8)}    & 60.0 \scriptsize{(0.9)} & 55.6 \scriptsize{(0.8)} & 60.5 \scriptsize{(1.1)} & 57.1 \scriptsize{(1.1)} \\

      \midrule
      \multirow{8}{*}{\rotatebox{90}{\cat}}

      & RACL-BERT & - & 67.5 & 70.3 & 52.4 & - & - & - & -\\
      \cmidrule(lr){2-10}

    & Head-first  & 43.0 \scriptsize{(1.3)} & 72.5 \scriptsize{(1.0)} & \impr{71.1 \scriptsize{(0.8)}} & 55.0 \scriptsize{(0.9)}    & \impr{66.8 \scriptsize{(0.5)}} & \impr{62.1 \scriptsize{(0.5)}} & \impr{62.0 \scriptsize{(1.1)}} & \impr{56.8 \scriptsize{(0.7)}} \\
    & +inlabel    & 43.1 \scriptsize{(2.2)} & \impr{73.4 \scriptsize{(1.0)}} & 70.3 \scriptsize{(1.0)} & \impr{55.8 \scriptsize{(1.8)}}    & 66.2 \scriptsize{(0.3)} & 61.5 \scriptsize{(0.6)} & 61.1 \scriptsize{(1.0)} & 56.0 \scriptsize{(1.0)} \\
    & Head-final  & 37.1 \scriptsize{(4.2)} & 71.2 \scriptsize{(0.6)} & 67.1 \scriptsize{(1.7)} & 53.9 \scriptsize{(2.2)}    & 62.7 \scriptsize{(0.4)} & 58.1 \scriptsize{(0.8)} & 59.7 \scriptsize{(1.1)} & 53.7 \scriptsize{(2.4)} \\
    & +inlabel    & 34.9 \scriptsize{(4.1)} & 70.7 \scriptsize{(1.4)} & 68.2 \scriptsize{(1.0)} & 53.5 \scriptsize{(0.7)}    & 63.4 \scriptsize{(0.5)} & 58.7 \scriptsize{(0.6)} & 60.9 \scriptsize{(1.1)} & 55.1 \scriptsize{(1.2)} \\
    & Dep. edges  & \impr{46.3 \scriptsize{(3.1)}} & 70.3 \scriptsize{(0.6)} & 69.2 \scriptsize{(1.4)} & 53.4 \scriptsize{(1.5)}    & 60.8 \scriptsize{(0.4)} & 57.5 \scriptsize{(0.6)} & 60.7 \scriptsize{(1.0)} & 55.6 \scriptsize{(0.9)} \\
    & Dep. labels & 45.6 \scriptsize{(2.9)} & 70.3 \scriptsize{(1.1)} & 69.1 \scriptsize{(1.7)} & 53.9 \scriptsize{(1.5)}    & 62.5 \scriptsize{(0.6)} & 59.1 \scriptsize{(0.6)} & 60.4 \scriptsize{(1.0)} & 55.8 \scriptsize{(1.2)} \\

      \midrule
      \multirow{8}{*}{\rotatebox{90}{\mpqa}}

      & RACL-BERT & - & 20.0 & 31.2 & 17.8 & - & - & - & -\\
      \cmidrule(lr){2-10}

    & Head-first  & 43.8 \scriptsize{(1.8)} & 51.0 \scriptsize{(1.9)} & \impr{48.1 \scriptsize{(0.8)}} & \impr{33.5 \scriptsize{(3.1)}}    & 40.0 \scriptsize{(1.0)} & 36.9 \scriptsize{(1.2)} & 24.5 \scriptsize{(2.3)} & 17.4 \scriptsize{(2.7)} \\
    & +inlabel    & 43.1 \scriptsize{(1.5)} & \impr{51.5 \scriptsize{(1.0)}} & 47.5 \scriptsize{(1.1)} & 21.3 \scriptsize{(0.4)}    & 40.6 \scriptsize{(0.5)} & 37.5 \scriptsize{(0.5)} & 24.5 \scriptsize{(1.3)} & 17.3 \scriptsize{(1.0)} \\
    & Head-final  & \impr{46.3 \scriptsize{(1.8)}} & 49.5 \scriptsize{(0.9)} & 46.0 \scriptsize{(1.1)} & 21.9 \scriptsize{(1.4)}    & \impr{41.4 \scriptsize{(0.7)}} & \impr{38.0 \scriptsize{(0.5)}} & \impr{26.1 \scriptsize{(0.7)}} & \impr{18.8 \scriptsize{(0.7)}} \\
    & +inlabel    & 45.6 \scriptsize{(2.5)} & 49.4 \scriptsize{(2.1)} & 45.6 \scriptsize{(1.1)} & 20.7 \scriptsize{(1.0)}    & 40.4 \scriptsize{(1.5)} & 37.2 \scriptsize{(1.9)} & 25.2 \scriptsize{(1.7)} & 17.8 \scriptsize{(1.3)} \\
    & Dep. edges  & 44.0 \scriptsize{(1.5)} & 48.5 \scriptsize{(1.2)} & 46.3 \scriptsize{(1.9)} & 18.9 \scriptsize{(2.3)}    & 35.4 \scriptsize{(1.3)} & 31.9 \scriptsize{(1.2)} & 24.2 \scriptsize{(1.6)} & 16.3 \scriptsize{(1.9)} \\
    & Dep. labels & 43.7 \scriptsize{(0.9)} & 47.7 \scriptsize{(2.3)} & 47.5 \scriptsize{(0.8)} & 21.9 \scriptsize{(0.7)}    & 35.6 \scriptsize{(1.2)} & 32.0 \scriptsize{(1.3)} & 24.0 \scriptsize{(0.8)} & 17.2 \scriptsize{(0.8)} \\

      \midrule
      \multirow{8}{*}{\rotatebox{90}{\dsu}}

      & RACL-BERT & - & 44.6 & 38.2 & 27.3 & - & - & - & -\\
      \cmidrule(lr){2-10}

        & Head-first & 28.0 \scriptsize{(7.7)}  & 39.9 \scriptsize{(2.2)} & 40.3 \scriptsize{(0.6)} & 26.7 \scriptsize{(2.1)}    & 35.3 \scriptsize{(0.9)} & 31.4 \scriptsize{(1.3)} & 31.0 \scriptsize{(1.4)} & 25.0 \scriptsize{(1.3)} \\
    & +inlabel   & 30.9 \scriptsize{(9.9)}  & 38.4 \scriptsize{(3.3)} & 40.6 \scriptsize{(2.9)} & 26.7 \scriptsize{(2.4)}    & 34.2 \scriptsize{(2.1)} & 30.7 \scriptsize{(2.5)} & 30.5 \scriptsize{(2.1)} & 25.4 \scriptsize{(2.3)} \\
    & Head-final & \impr{37.4 \scriptsize{(11.6)}} & \impr{42.1 \scriptsize{(2.7)}} & \impr{45.5 \scriptsize{(2.4)}} & \impr{29.6 \scriptsize{(1.7)}}    & \impr{38.1 \scriptsize{(1.9)}} & \impr{33.9 \scriptsize{(2.3)}} & \impr{34.3 \scriptsize{(4.2)}} & 26.5 \scriptsize{(3.5)} \\
    & +inlabel   & 30.6 \scriptsize{(16.4)} & 38.9 \scriptsize{(3.1)} & 45.2 \scriptsize{(2.7)} & 28.1 \scriptsize{(3.7)}    & 37.3 \scriptsize{(2.7)} & 33.3 \scriptsize{(2.1)} & 29.4 \scriptsize{(2.8)} & 23.7 \scriptsize{(2.4)} \\
        & Dep. edges  & 32.7 \scriptsize{(12.1)} & 39.9 \scriptsize{(2.8)} & 44.8 \scriptsize{(4.0)} & 28.9 \scriptsize{(3.6)}    & 37.3 \scriptsize{(2.5)} & 33.8 \scriptsize{(2.7)} & 33.2 \scriptsize{(4.5)} & \impr{27.3 \scriptsize{(4.1)}} \\
    & Dep. labels & 30.8 \scriptsize{(5.8)}  & 38.9 \scriptsize{(0.9)} & 43.1 \scriptsize{(1.2)} & 27.8 \scriptsize{(1.9)}    & 35.7 \scriptsize{(1.5)} & 32.1 \scriptsize{(1.6)} & 31.3 \scriptsize{(1.1)} & 25.3 \scriptsize{(2.0)} \\
    \bottomrule
    \end{tabular}
    }
    \caption{Experiments with mBERT.}
    \label{tab:all_others}
\end{table*}

\begin{table}[]
\begin{tabular}{@{}ll@{}}
\hline
    \textbf{GPU Infrastructure} & NVIDIA P100, 16 GiB RAM \\
\hline
    \textbf{CPU Infrastructure} & Intel Xeon-Gold 6138 2.0 GHz \\
\hline
    \textbf{Training duration} & 00:31:43 (\basque) -- 07:40:54 (\norec) \\
\hline
    \textbf{Model implementation} & \url{https://github.com/jerbarnes/sentiment_graphs/src} \\
\hline
    \textbf{Hyperparameter}  & \textbf{Best assignment} \\
\hline
    embedding & Word2Vec SkipGram 100D \\
\hline
contexualized embedding & mBERT \\
\hline
    embeddings trainable & False \\
\hline
    number of epochs & 100  \\
\hline
    batch size & 50  \\
\hline
beta1  & 0\\
\hline
beta2 & 0.95\\
\hline
l2 & 3e-09\\
\hline
hidden lstm & 200\\
\hline
hidden char lstm & 100\\
\hline
layers lstm & 3\\
\hline
dim mlp & 200\\
\hline
dim embedding & 100\\
\hline
dim char embedding & 80\\
\hline
early stopping & 0\\
\hline
pos style & xpos\\
\hline
attention & bilinear\\
\hline
model interpolation & 0.5\\
\hline
loss interpolation & 0.025\\
\hline
lstm implementation & drop connect\\
\hline
char implementation & convolved\\
\hline
emb dropout type & replace\\
\hline
bridge & dpa+\\
\hline
dropout embedding & 0.2\\
\hline
dropout edge & 0.2\\
\hline
dropout label & 0.3\\
\hline
dropout main recurrent & 0.2\\
\hline
dropout recurrent char & 0.3\\
\hline
dropout main ff & 0.4\\
\hline
dropout char ff & 0.3\\
\hline
dropout char linear & 0.3\\
\hline
\end{tabular}
\end{table}

\begin{figure}[h!]
    \centering
    \includegraphics[width=.45\textwidth]{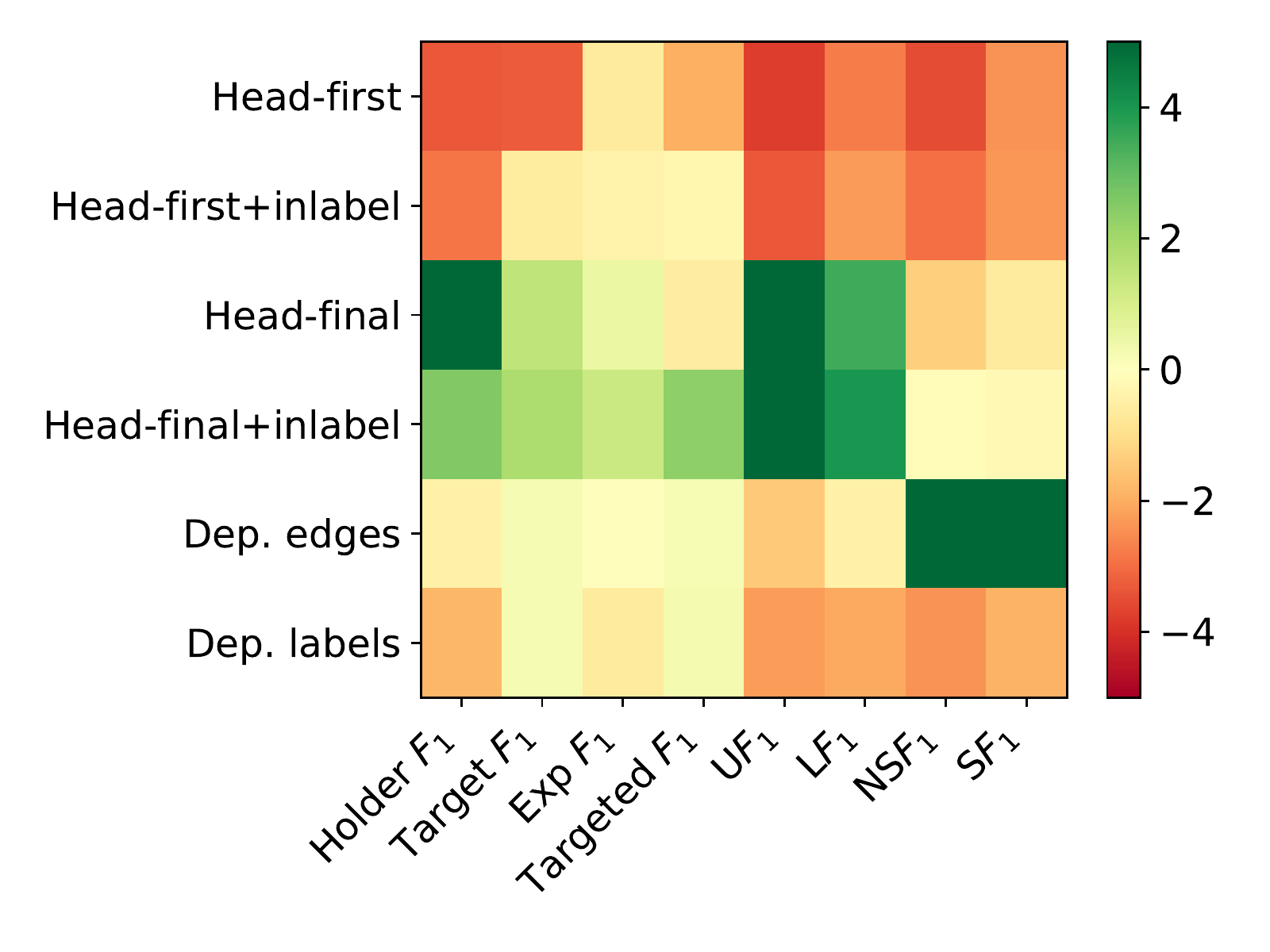}
    \caption{Average benefit of each graph annotation scheme (y-axis) on the evaluation metrics (x-axis) for \norec.}
    \label{fig:aggregate_norec}
\end{figure}

\begin{figure}[h!]
    \centering
    \includegraphics[width=.45\textwidth]{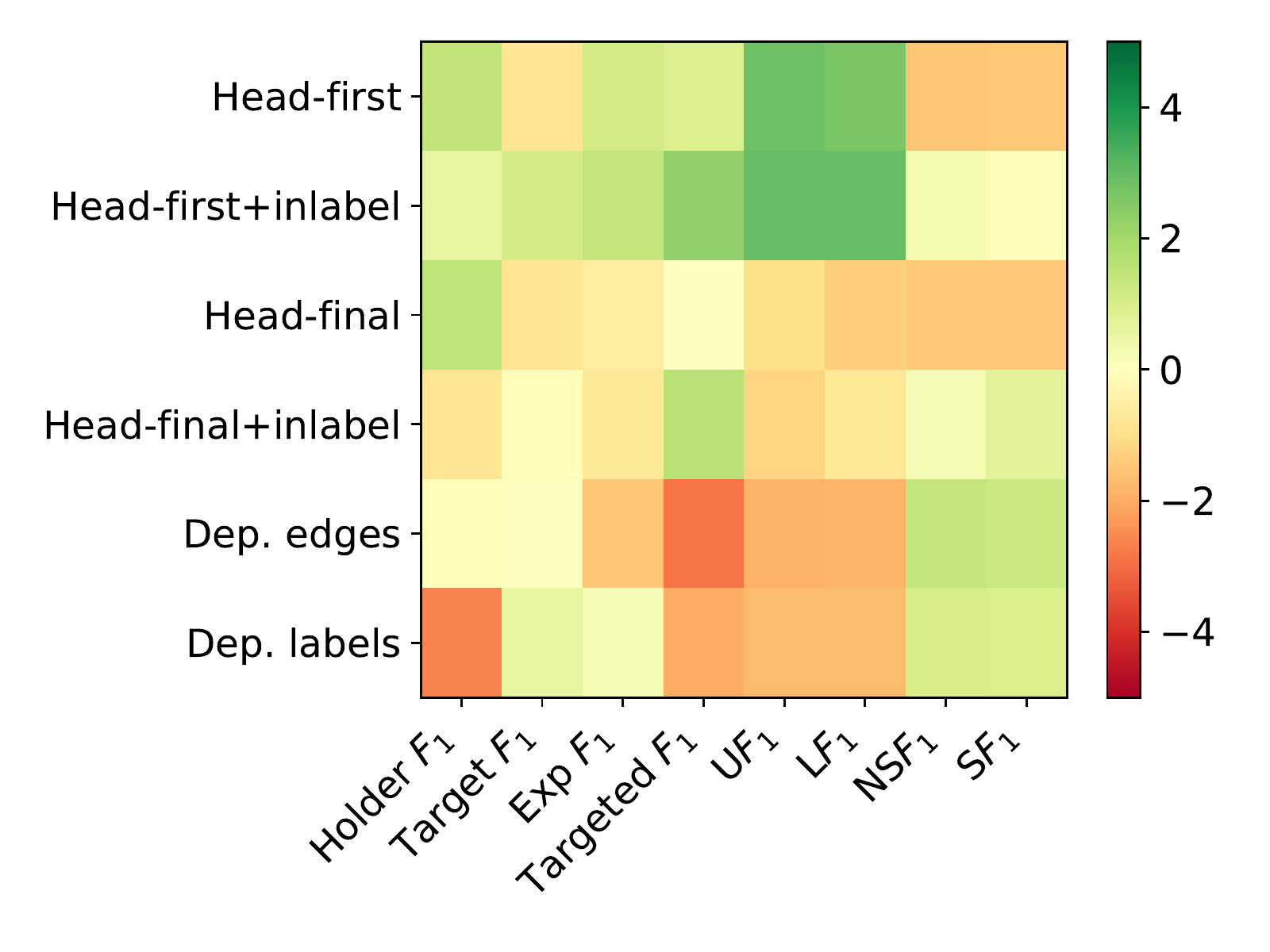}
    \caption{Average benefit of each graph annotation scheme (y-axis) on the evaluation metrics (x-axis) for \basque.}
    \label{fig:aggregate_eu}
\end{figure}

\begin{figure}[h!]
    \centering
    \includegraphics[width=.45\textwidth]{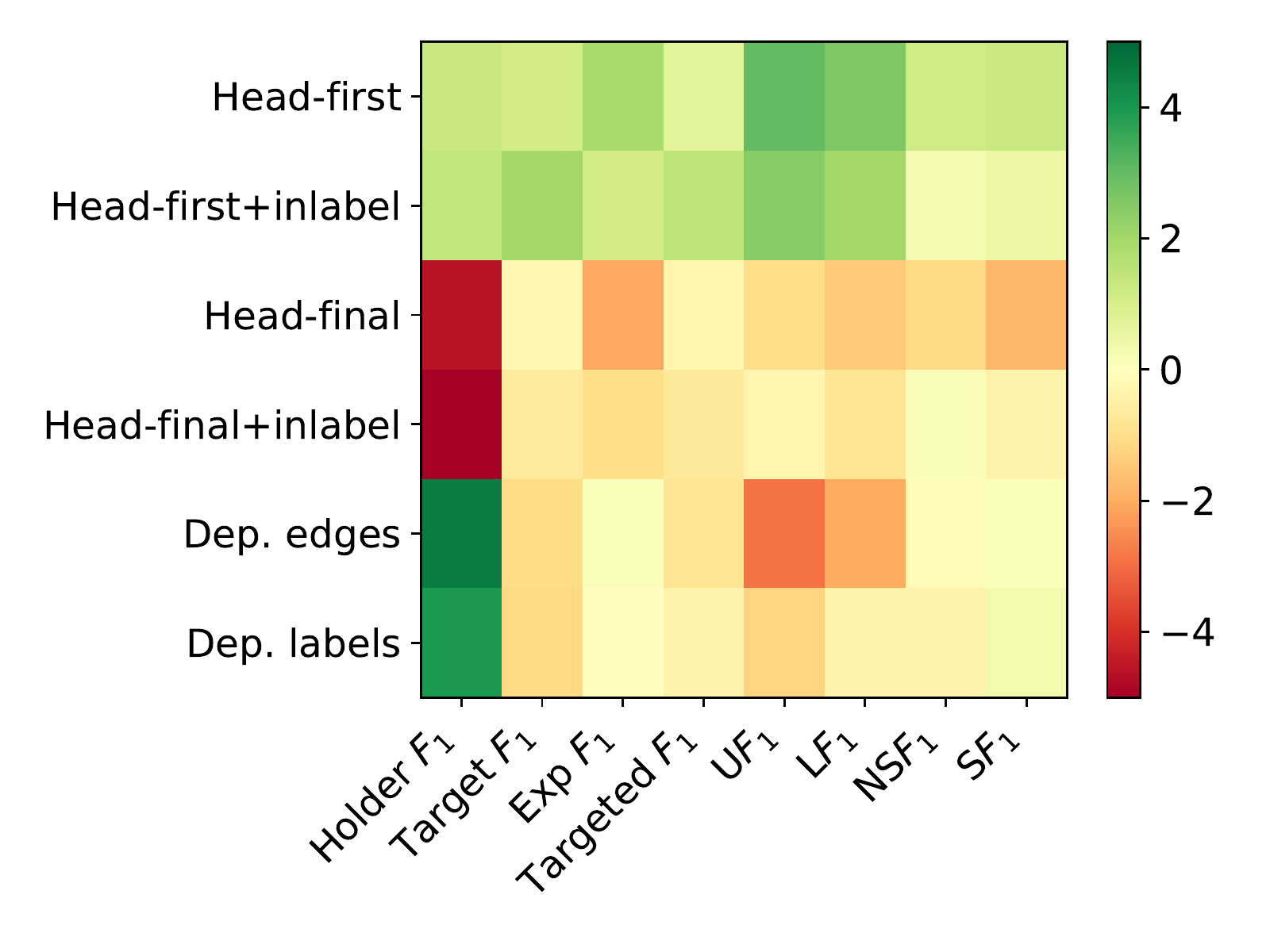}
    \caption{Average benefit of each graph annotation scheme (y-axis) on the evaluation metrics (x-axis) in percentage points for \cat.}
    \label{fig:aggregate_cat}
\end{figure}

\begin{figure}[h!]
    \centering
    \includegraphics[width=.45\textwidth]{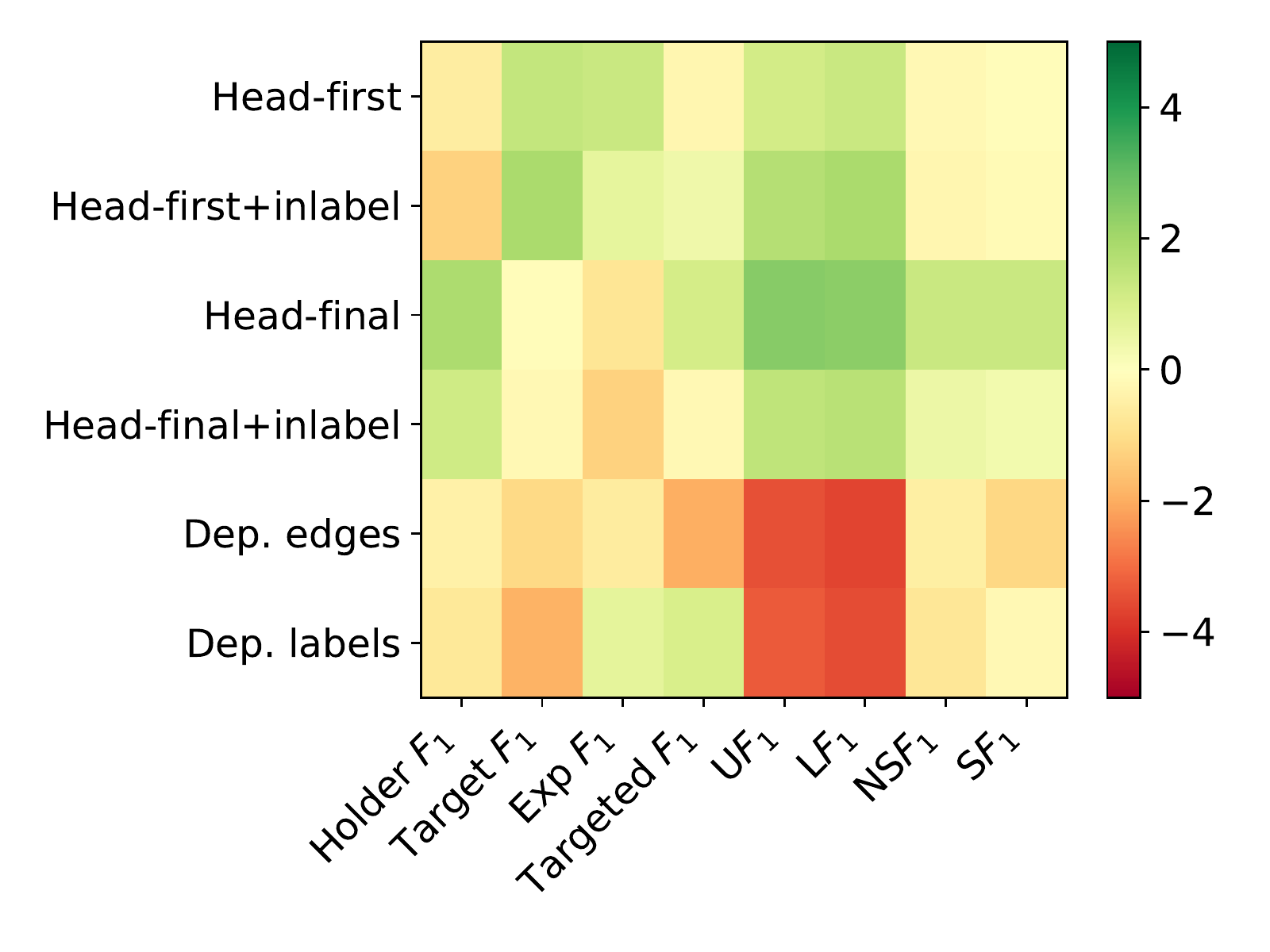}
    \caption{Average benefit of each graph annotation scheme (y-axis) on the evaluation metrics (x-axis) in percentage points for \mpqa.}
    \label{fig:aggregate_mpqa}
\end{figure}

\begin{figure}[h!]
    \centering
    \includegraphics[width=.45\textwidth]{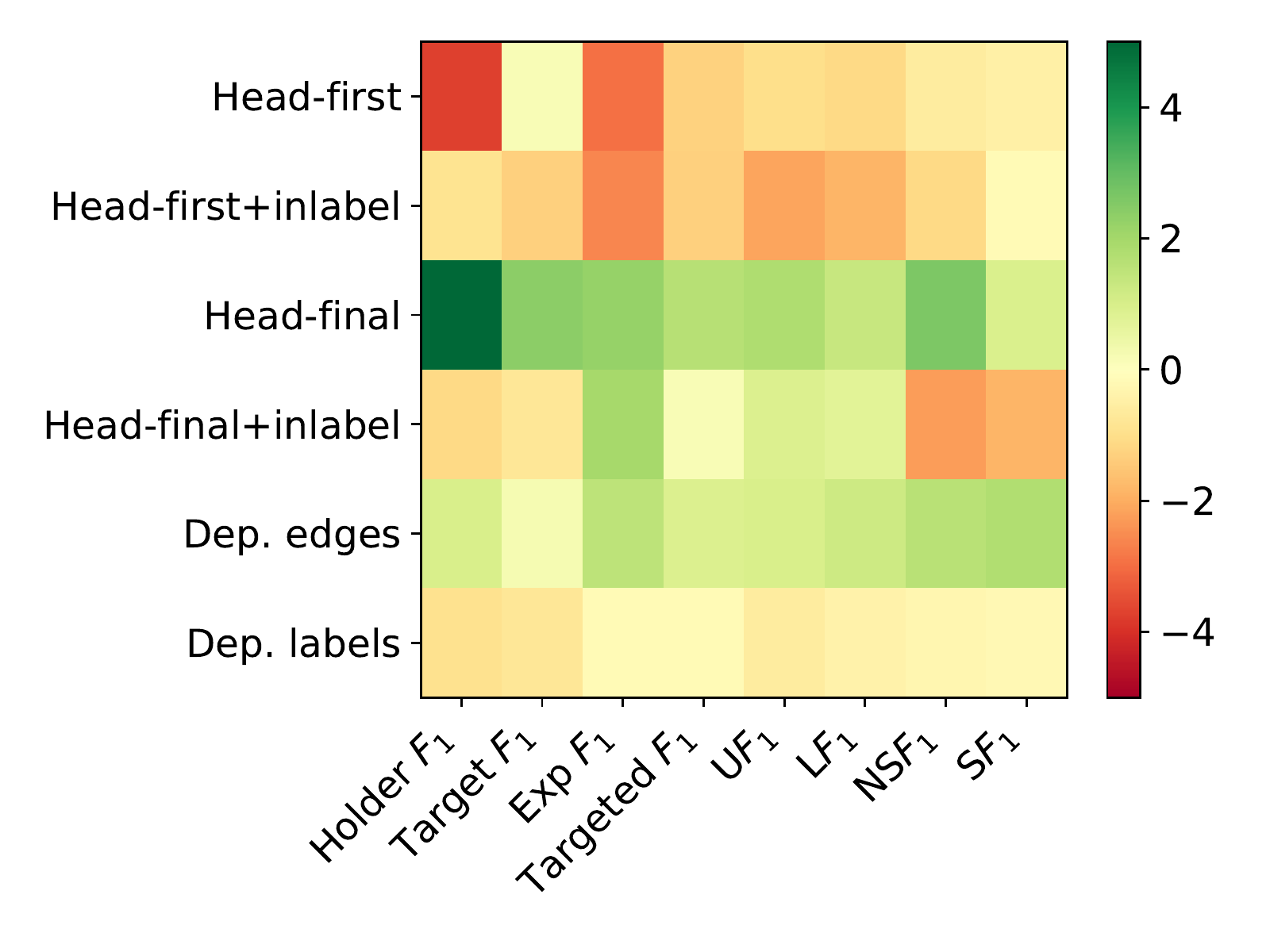}
    \caption{Average benefit of each graph annotation scheme (y-axis) on the evaluation metrics (x-axis in percentage points) for \dsu.}
    \label{fig:aggregate_dsunis}
\end{figure}

\begin{figure}[h!]
    \centering
    \includegraphics[width=.45\textwidth]{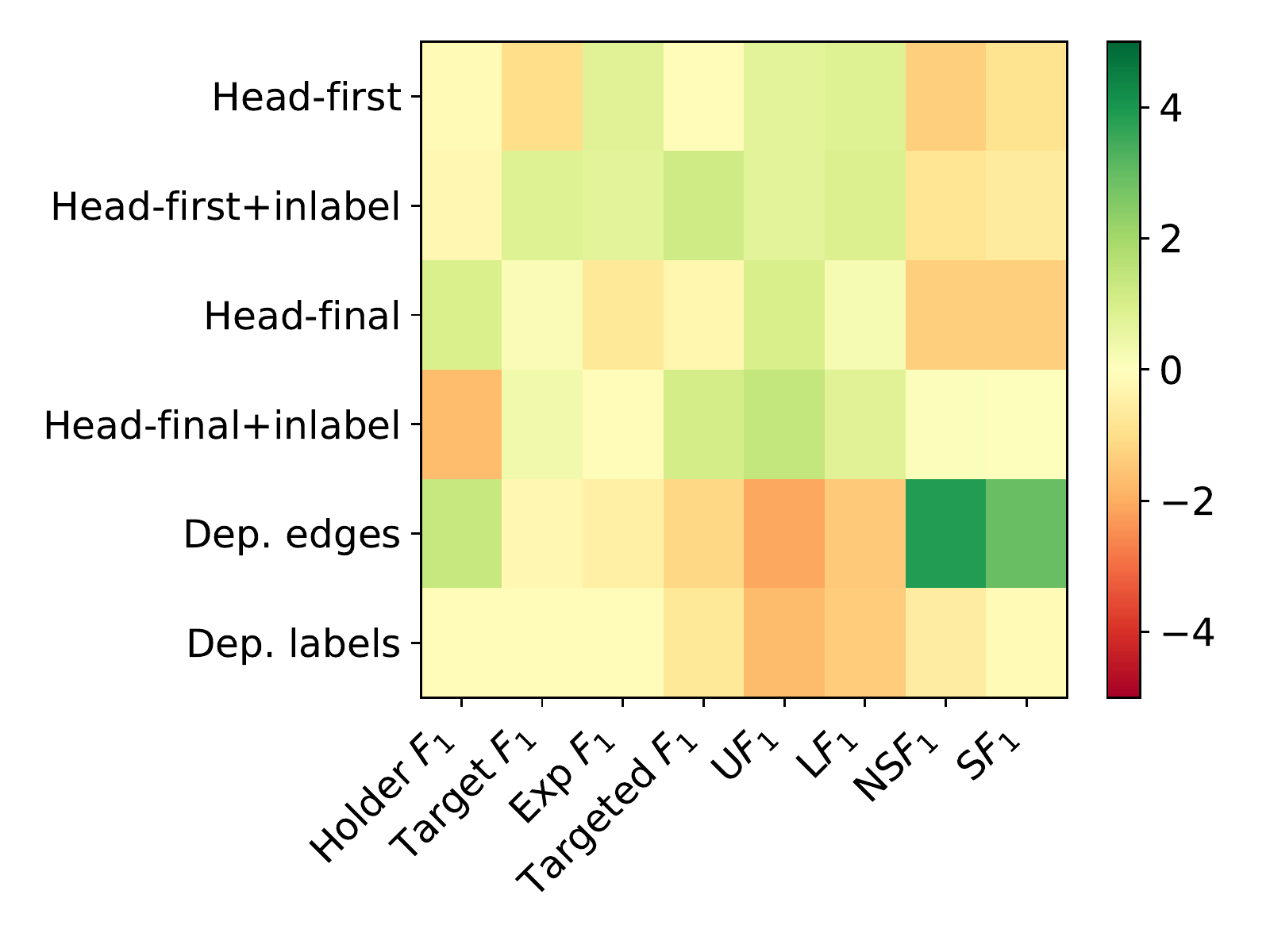}
    \caption{Average benefit of each graph annotation scheme (y-axis) on the evaluation metrics (x-axis) in percentage points. The results on \norec, \basque, \cat.}
    \label{fig:aggregate3}
\end{figure}

\end{document}